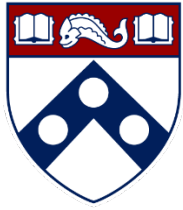

# Generative Interpretation

**Yonathan A. Arbel**
*University of Alabama - School of Law*

**David A. Hoffman**
*University of Pennsylvania Carey Law School*

# Generative Interpretation

*Yonathan Arbel & David A. Hoffman*[*]

[*DRAFT* July 30, 2023]

*We introduce generative interpretation, a new approach to estimating contractual meaning using large language models. As AI triumphalism is the order of the day, we proceed by way of grounded case studies, each illustrating the capabilities of these novel tools in distinct ways. Taking well-known contracts opinions, and sourcing the actual agreements that they adjudicated, we show that AI models can help factfinders ascertain ordinary meaning in context, quantify ambiguity, and fill gaps in parties' agreements. We also illustrate how models can calculate the probative value of individual pieces of extrinsic evidence.*

*After offering best practices for the use of these models given their limitations, we consider their implications for judicial practice and contract theory. Using LLMs permits courts to estimate what the parties intended cheaply and accurately, and as such generative interpretation unsettles the current interpretative stalemate. Their use responds to efficiency-minded textualists and justice-oriented contextualists, who argue about whether parties will prefer cost and certainty or accuracy and fairness. Parties—and courts—would prefer a middle path, in which adjudicators strive to predict what the contract really meant, admitting just enough context to approximate reality while avoiding unguided and biased assimilation of evidence. As generative interpretation offers this possibility, we argue it can become the new workhorse of contractual interpretation.*

---

[*] Associate Professor, University of Alabama Law & William A. Schnader Professor, University of Pennsylvania Carey School of Law. We thank Michael Hurley, Elizabeth Meeker and JD Uglum for helpful research assistance.

## *TABLE OF CONTENTS*

# INTRODUCTION

When New Orleans' levees broke during Hurricane Katrina, devastation, both human and economic, swept the city. And then came the lawyers. In mass contract litigation by policyholders against their insurance companies, advocates fighting over tens of billions of dollars of potential liability ultimately contested the meaning of a single word, representing a concept the companies had excluded from coverage. *Flood*.[1] Plaintiffs labored first to convince judges that *flood* might not mean water damage caused by humans, so they could then prove to a fact-finder that their insurance policies didn't contemplate damage resulting from negligence by the Army's Corps of Engineers.[2] Lawyers for the defense argued that the word was unambiguous, covering rising waters no matter their cause, and therefore no further factfinding was necessary.[3] Here, as so often in real court proceedings, though rarely in law school classrooms, expensive, cumbersome and unsatisfactory processes of contract interpretation took center stage.[4]

---

[1] In re Katrina Canal Breaches Litig., 495 F.3d 191, 199 (5th Cir. 2007) ("We will not pay for loss or damage caused directly or indirectly by any of the following. Such loss is excluded regardless of any other cause or event contributing concurrently or in any sequence to the loss. .... Water ... Flood, surface water, waves, tides, tidal waves, overflow of any body of water, or their spray, all whether driven by wind or not ....").

[2] In re Katrina Canal Breaches Litig., 495 F.3d 191, 197, 199, 200-01, 203-04 (5th Cir. 2007); Brief for Appellee-Cross Appellant Humphreys at 16-18, In re Katrina Canal Breaches Litig., 495 F.3d 191 (5th Cir. 2007) (No. 07-30119), 2007 WL 4266576; Brief for Plaintiff-Appellee Xavier Univ. of La. at 17-44, In re Katrina Canal Breaches Litig., 495 F.3d 191 (5th Cir. 2007) (No. 07-30119), 2007 WL 4266583; Brief of the Chehardy Representative Policyholders in Response at 14-41, In re Katrina Canal Breaches Litig., 495 F.3d 191 (5th Cir. 2007) (No. 07-30119), 2007 WL 4266578. On the scope, source, and allocation of negligence see ANDY HOROWITZ, KATRINA: A HISTORY, 1915-2015, 1-12, 128-33 (2020); *see also* Campbell Robertson & John Schwartz, *Decade After Katrina, Pointing Finger More Firmly at Army Corps*, THE NEW YORK TIMES, May 23, 2015, https://www.nytimes.com/2015/05/24/us/decade-after-katrina-pointing-finger-more-firmly-at-army-corps.html.

[3] In re Katrina Canal Breaches Litig., 495 F.3d at 208. Brief of Appellee State Farm Fire & Casualty Co. at 14-26, In re Katrina Canal Breaches Litig., 495 F.3d 191 (5th Cir. 2007) (No. 07-30119), 2007 WL 2466572; Brief of Appellee Allstate Ins. Co. & Allstate Indem. Co. at 16-37, In re Katrina Canal Breaches Litig., 495 F.3d 191 (5th Cir. 2007) (No. 07-30119), 2007 WL 4266556.

[4] Benjamin E. Hermalin, Avery W. Katz & Richard Craswell, *Contract Law, in* 1 HANDBOOK OF LAW AND ECONOMICS 3, 68 (A. Mitchell Polinsky & Steven Shavell eds., 2007) (noting that interpretation is the most litigated type of contract dispute).

After years of litigation, the Fifth Circuit—in the best-known and most conse-quential contracts case of the last generation[5]—held that *flood* was unambiguous: it meant any inundation, regardless of cause.[6] To get to that outcome, it engaged in the most artisanal and articulated form of textualism available in late-stage Capitalism. The court consulted four dictionaries, one encyclopedia, two treatises, a medley of for-and-against in-and-out-of-jurisdiction cases, and two linguistic, latinized interpretative can-ons.[7] That's on top of the four dictionaries and twenty reporter pages of caselaw analyz-ing the same problem in the district court.[8]

Notwithstanding such expensive and extensive efforts, the court's interpretation has come under attack: its dictionary analysis was misleading,[9] its canons badly de-ployed,[10] and some of the relevant legal authorities were in fact pro-plaintiff.[11] Rather than reach a decision that followed from a constraining method, the Fifth Circuit (says its critics) merely affirmed its pro-business priors.[12] If textualism looks like another infinitely malleable and justificatory practice in high stakes cases, what good is it? But textualism's competitor, kitchen-sink contextualism, has been in bad odor for two generations, at least for the sorts of contracts that generally get litigated.[13] Thus, contract jurists muddle along, looking for a better, more convenient path.[14]

---

[5] The opinion has been cited nearly 7,000 times over 15 years, discussed in almost 2,000 secondary sources, and is taught to 1Ls. *See, e.g.*, IAN S. AYRES AND GREGORY M. KLASS, STUDIES IN CON-TRACT LAW 701 (9ᵀᴴ ED. 2017).

[6] In re Katrina Canal Breaches Litig., 495 F.3d at 214-19 ("The distinction between natural and non-natural causes in this context would . . . lead to absurd results and would essentially eviscerate flood exclusions whenever a levee is involved.").

[7] *Id.* at 210-19.

[8] In re Katrina Canal Breaches Consolidated Litig., 466 F.Supp.2d 729, 747-763 (E.D.La. 2006).

[9] Natasha Fossett, *What Does Flood Mean to You: The Louisiana Courts' Struggle to Define in Sher v. Lafayette Insurance Company,* 37 S.U. L. REV. 289, 303-306 (2010) (arguing that flood as defined in Louisiana Law had a narrower meaning than either the Fifth Circuit or the later Louisiana Supreme Court decision implied).

[10] Rachel Lisotta, *In Over Our Heads: The Inefficiencies of the National Flood Insurance Program and the Institution of Federal Tax Incentives,* 10 LOY. MAR. L. J. 511, 523 (2012) (criticizing the court for not focusing on the intent of the parties); Fossett, *supra* note 9, at 309-310 (arguing for use of the absurdity canon). Mark R. Patterson, *Standardization of Standard-Form Contracts: Competition and Contract Implications,* 52 WM. & MARY L. REV. 327, 356 (2010) (critiquing the Fifth Circuit for failing to address the significance of the relevant policy being drafted by the Insurance Service Office); Eyal Zamir, *Contract Law and Theory: Three Views of the Cathedral,* 81 U. CHI. L. REV. 2077, 2096 (2014) (critiquing the limited tools used by American courts to regulate standard form contracts, as evidenced by the court's narrow approach in the Katrina case).

[11] *See, e.g.,* Sher v. Lafayette Ins. Co., 2007-CA-0757, 2007 WL 4247708 (La. App. 4th Cir. Nov. 19, 2001) (finding flood ambiguous), reversed by Sher v. Lafayette Ins. Co., 07-2441 (La. 4/8/08); 988 So. 2d 186; Ebbing v. State Farm Fire & Cas. Co., 1 S.W.3d 459, 462 (Ark. Ct. App. 1999) (holding flood excluded manmade causes); *cf.* M & M Corp. of S.C. v. Auto-Owners Ins. Co., 701 S.E.2d 33 (S.C. 2010) (finding that rainwater deliberately channeled on insured's land was not flood water).

In this article we offer a new approach to determining contracting parties' meaning, which we'll call *generative interpretation*.[15] The idea is simple: applying large language models (LLMs) to contractual texts and extrinsic evidence to predict what the parties would have said at contracting about what they meant.[16] Our goal is to convince you that generative interpretation avoids some of the problems that bedeviled the Fifth Circuit in its Katrina litigation, while being materially more accessible and transparent. Giving courts a convenient way to commit to a cheap and predictable contract interpretation methodology would be a major advance in contract law, and we argue that even today's freshly-minted LLMs can be of service.

---

[12] Willy E. Rice, *The Court of Appeals for the Fifth Circuit: A Review of 2007-2008 Insurance Decisions,* 41 TEX. TECH L. REV. 1013, 1039 (2009) ("[T]he Fifth Circuit has received some highly negative coverage in newspapers for its pro-insurer, Katrina-related decisions . . . Without doubt, for those who believe the Fifth Circuit is a "pro-insurer court," the discussions of the outcomes and opinions in those cases will do very little to dispel that perception."); Kenneth S. Abraham & Tom Baker, *What History Can Tell Us About the Future of Insurance and Litigation After Covid-19,* 71 DEPAUL L. REV. 169, 189 (2022) (arguing that homeowners' unwillingness to buy federal flood insurance helped motivate strict construction of their private contracts); Thomas A. McCann, *5th Circuit Ruling: A Tough Pill to Swallow for Katrina Policyholders,* 20 LOY. CONSUMER L. REV. 100 (2007); Becky Yerak, *Insurers Win Key Katrina Ruling,* CHICAGO TRIBUNE, Aug. 3, 2007, at C1 (noting the effect on homeowners). To be clear, the earlier ruling came under even more scrutiny. *See, e.g.* Walter J. Andrews, Michael S. Levine, Rhett E. Petcher, and Steven W. McNutt, Essay, *A "Flood of Uncertainty": Contractual Erosion in the Wake of Hurricane Katrina and the Eastern District of Louisiana's Ruling in In Re Katrina Canal Breaches Consolidated Litigation,* 81 TUL. L. REV. 1277 (2006) (arguing that the District Court's finding that flood was ambiguous was wrong); Michelle E. Boardman, *The Unpredictability of Insurance Interpretation,* 82 L. & CONTEMP. PROBS. 27, 41 n.45 (2019) (calling the District Court infamous and arguing that the Fifth Circuit ruling was correct); Edward P. Richards, *The Hurricane Katrina Levee Breach Litigation: Getting the First Geoengineering Liability Case Right,* 160 U. PA. L. REV. 267 (2012) (arguing in support of the Fifth Circuit ruling).

[13] Lawrence A. Cunningham, *Contract Interpretation 2.0: Not Winner-Take-All but Best-Tool-For-The-Job,* 85 GEO. WASH. U. L. REV. 1625, 1628-31 (offering the history of contextualism versus textualism and noting a rise in the latter starting in the early 1990s); *but cf.* 5 CORBIN ON CONTRACTS § 24.7 (2023) (noting a "trend" toward abandoning plain meaning in some states).

[14] Cunningham, *supra,* at 1633-1643 (noting proposals to compromise between the two approaches).

[15] For previous discussions of the use of large language models in contracts, see Ryan Catterwell, *Automation in Contract Interpretation,* 12 L. INNOVATION & TECH. 81, 100 (2020) (early paper showing how information can be extracted from contractual texts); Yonathan A. Arbel & Shmuel I. Becher, *Contracts in the Age of Smart Readers,* 90 GEO. WASH. L. REV. 83 (2022) (arguing that language models could serve as "smart readers" of consumer contracts); Noam Kolt, *Predicting Consumer Contracts,* 37 BERK. TECH. L.J. 71 (2022) (arguing that ChatGPT might be useful in helping consumers to understand their contracts and providing examples).

Convincing judges to forgo dictionaries and canons and adopt a chat tool best known today for encouraging lawyers to submit fake authorities will be a tall order.[17] We'll largely proceed by way of demonstrative case studies. Let's start with the word *flood*. In the Katrina case, the question was really whether the widely shared meaning of *flood* reasonably excluded manmade disasters. To answer that question you could, as the court did, turn to the traditional tools of High Textualism. Or you could survey insured citizens (if you could identify them and avoid motivated answers).[18] And you might even, if you were technically sophisticated and patient enough, query a few relatively small databases and ask which words in English generally tend to occur, or collocate, with flood in newspapers, books, and the like.[19]

But we instead turned to a convenient, free, open-source LLM tool resting on a database of trillions of words and asked it to transform words into complex vectors in a process called *embedding*.[20] As a first cut, this process can be thought of as trying to quantify how much a word belongs to a given category, or dimension. Thus, if there is a dimension for the word *water*, *fish* will score higher than *dogs*. Using an interface we developed, we queried several models about the relation of word flood in its contractual context (attributing flood to water damage) to other potential environments.[21]

---

[16] *Cf.* Jonathan H. Choi, *Measuring Clarity in Legal Texts,* 91 U. CHI. L. REV. (forthcoming, 2024). Choi's excellent paper, though not focused on contract interpretation particularly, significantly advanced understanding of how automated interpretative methods can aid factfinders. We build on his work technically by developing new ways of interacting with large language models and incorporating context and attention mechanisms.

[17] *See infra* at text accompanying notes 181 to 184 (discussing Mata v. Avianca, Inc., __ F.Supp.3d. __, 2023 WL 4114965 (June 22, 2023).); *see also* Ex Parte Allen Michael Lee, __. S.W.3d. __ , 2023 WL 4624777, at *1 n.2 (Ct. App. Tex. July 19, 2023) (explaining the court's suspicion that counsel had filed briefs using ChatGPT and had made up cases and citations).

[18] *See* Omri Ben-Shahar & Lior J. Strahilevitz, *Interpreting Contracts via Surveys and Experiments*, 92 N.Y.U. L. REV. 1753 (2017) (proposing using surveys to interpret certain mass contracts).

[19] *See* Stephen C. Mouritsen, *Contract Interpretation with Corpus Linguistics*, 94 WASH. L. REV. 1337, 1378 (2019) (proposing using corpus linguistics to interpret contracts).

[20] For a survey of embedding methods, see MOHAMMAD TAHER PILEHVAR & JOSE CAMACHO-COLLADOS, EMBEDDINGS IN NATURAL LANGUAGE PROCESSING 27-110 (2021).

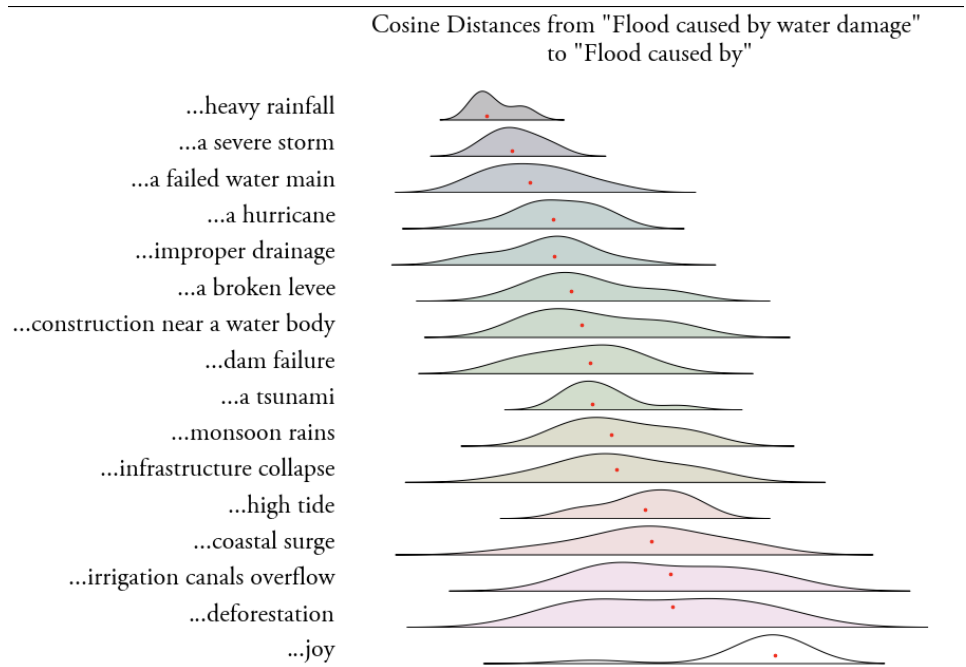

*Figure 1: Analysis of the cosine distance—a measure of distance for the numerical representation of terms (embeddings) by language models—between an attention-weighted clause on the cause of flood and other terms.*

To read Figure 1, focus on the location of the red markers. The farther they are from the origin, the more distant the term is from the *flood* clause. In our view, the Figure offers immediately available, objective, cheap support for the court's judgment that floods can be unnaturally caused. Common sentences regarding floods do not distinguish between the type of cause, but rather seem more focused on its typicality. Our quality checks, *Flood caused by joy* or *deforestation*, are indeed farther out than flood caused by *rainfall* or a *storm*. And while it supports this decision of the court, it challenges another. Louisiana courts refused to exclude water main floods, even though linguistically they appear to be as much of a flooding event as any other.[22]

---

[21] All of the code necessary to replicate these results, and the remaining ones in the paper, can be found at: https://github.com/yonathanarbel/generativeinterpretation/tree/main. Because embeddings are vectors in high-dimensional space, we can measure the distance between them. This method has been used extensively in the literature. *See* Choi, *supra* note 16, 24-26 (using method and reports its usage and limitations.) For a non-legal example, *see e.g.*, Nitika Mathur, Timothy Baldwin & Trevor Cohn, *Putting Evaluation in Context: Contextual Embeddings Improve Machine Translation Evaluation*, PROCEEDINGS OF THE 57TH ANNUAL MEETING OF THE ASSOCIATION FOR COMPUTATIONAL LINGUISTICS 2799 (2019). We found that while results using this method seem sensible, they are also fragile. To create a more robust measure, we relied on the embeddings of the ten top performing models today (found at https://huggingface.co/spaces/mteb/leaderboard on pair classification tasks) and used similar sentence structures. This approach is partly inspired by Maria Antoniak & David Mimno, *Evaluating the Stability of Embedding-based Word Similarities*, 6 TRANS. ASS'N FOR COMPUTATIONAL LINGUISTICS 107 (2018). We then calculated the cosine distance, normalized it, and reported the results in the figure below.

Now, the model doesn't provide (nor could it) a scientific way to judge whether words are sufficiently close to make the plain meaning of *flood* unambiguous. But there is a bit of difference between an informed conclusion based on a statistical analysis of billions of texts and a judgment by a few dictionary editors. And there is an ocean of difference between the baroque and expensive textualism the court used and code that is cheap, replicable, quick, and most importantly, extremely straightforward to use. Simply put, generative interpretation is good enough for many cases that currently employ more expensive, and arguably less certain, methodologies. It's a workable, workmanlike method for a resource-constrained contract litigation world.

We introduce the methodologies of contract interpretation (in Part I) and argue that they badly fail at their core purposes of unbiased, accessible ascertainment of what the parties would have wanted. In practice, interpretation operates as a kludgy prediction engine. Both textualism and contextualism strive to estimate what the parties would have said on a matter, accounting for realistic constraints of evidence and cost. But those constraints impose real tradeoffs and can't avoid legitimacy problems generated by courts' motivated reasoning. We describe some modern proposed improvements on interpretation's normal science and suggest that however promising they are, concerns about usability and cost impair their real-world utility.[23]

Part II is the heart of the Article. Here, we look at several types of interpretative problems generated by real contracts that produced contracts opinions. These range from the easy (the predicted meaning of a particular word) to the hard (whether there is an ambiguity) to the metaphysical (what did the parties mean when they clearly hadn't considered the issue). In each example, we showcase new ways to use large language models to sharpen intuitions about the parties' presumed intent, to illuminate how transparent and objective interpretative methodologies have advantages over intuitive ones, and to suggest that generative interpretation has real promise as a judicial adjunct. The cases we run through include casebook staples, like *Trident Ctr. v. Connecticut Gen. Life Ins. Co.*[24] and *C & J Fertilizer, Inc. v. Allied Mut. Ins. Co.*,[25] as well as some that should be, like *Famiglio v. Famiglio*,[26] *Haines v. City of New York*[27] and *Stewart v. Newbury*.[28] For many of these cases, our work is based on archival research identifying original contract materials, until now obscured by the judicial opinions that purportedly interpret them.

---

[22] Sher v. Lafayette Ins. Co., 2007-2441 (La. 4/8/08), 988 So. 2d 186, 195, on reh'g in part (July 7, 2008) ("inundation of property due to broken water mains . . . would not be excluded as a 'flood'"). *In re Katrina Canal Breaches Litig.*, 495 F.3d at 216 ("Unlike a canal, a water main is not a body of water or watercourse.").

[23] *See infra* at text accompanying notes 32 to 107.

These case studies show how generative interpretation might be deployed in practice. As we will explore, the technology underlying large language models can do more than merely helping us to see if *flood* is closer to *levee* than it is to *joy*. Dictionaries, encyclopedias, or corpus linguistics can do that. What makes large language models powerful is the vastness of the data they incorporate; what makes them unique is that they wield an internal mechanism known as "attention" which allows them to attend to context. And by becoming context sensitive, these models can parse out the effects of contract text from the marginal value of relevant extrinsic evidence

But current practices about LLMs and their future uses are contingent: lawyers tend to use tools before they are theoretically sharp.[29] In Part III, we develop a theory to justify and constrain generative interpretation going forward, as the technology that enables it continues to rapidly develop and its use by lawyers and judges grows explosively. We make two claims.

---

[24] 847 F.2d 564 (9th Cir. 1988). *See, e.g.*, RANDY E. BARNETT & NATHAN B. OMAN, CONTRACTS: CASES AND DOCTRINE 483 (7th ed. 2021); E. ALLEN FARNSWORTH, CAROL SANGER, NEIL B. COHEN, RICHARD R.W. BROOKS AND LARRY T. GARVIN, CASES AND MATERIALS ON CONTRACTS __ (10th ed. 2023).

[25] 227 N.W.2d 169 (Iowa 1975). *See* Brian Bix, *The Role of Contract: Stewart Macaulay's Lessons from Practice*, in REVISITING THE CONTRACTS SCHOLARSHIP OF STEWART MACAULAY: ON THE EMPIRICAL AND THE LYRICAL 252 (Jean Braucher, John Kidwell & William Whitford eds., Hart Publishing, 2013) (describing C&J and noting that it is often assigned in casebooks, including Stewart Macaulay's and Charles Knapp's).

[26] 279 So.3d 736 (Fla. Dist. Ct. App. 2019).

[27] 41 N.Y.2d 769 (1977). ROBERT S. SUMMERS, ROBERT A. HILLMAN AND DAVID A. HOFFMAN, CONTRACT AND RELATED OBLIGATION: THEORY, DOCTRINE, AND PRACTICE 834 (8th ed. 2021).

[28] 220 N.Y. 379 (1917). SUMMERS ET AL., *supra*, at 948.

[29] Consider originalism.

*First*, the method fills a glaring need for a simple, transparent and convenient way to commit to an interpretative method which helps predict the parties' intent. If courts follow the set of best practices we describe, they will avoid certain access-to-justice and legitimacy problems that have beset the modern contract litigation machine. *Second*, rather than simply a marginal improvement over dictionary-and-canon textualism, or its negation as a form of 1960s-California contextualism,[30] use of artificial intelligence (AI) should prompt a reexamination of the utility of those categories entirely. As more courts commit to generative interpretation, parties may come to prefer contextual evaluation of meaning when their deals are evaluated, thus flipping a longstanding default rule in contract law.[31]

We do consider some of the developing objections to the use of large language models, including their hallucinatory errors, biases, black-box methods, and the tension between the rapidity of their deployment and stately needs of precedential decision-making. As we show, generative interpretation's dangers illustrate its limits: judges will have to use these engines as *tools* to excavate the normative judgments on which all interpretative and adjudicatory exercises rest. Large language models aren't robot judges. What they will do (and maybe are already doing) is help judges illuminate the degree to which we want to give the parties what they really bargained for, as best as we can.

## I.    CONTRACT INTERPRETATION AS PREDICTION

Jurists interpreting contracts start with a simple question: "what would the parties have said about the meaning of a disputed phrase at the time they entered the contract?"[32] That is, to "ascertain the parties' intention at the time [the parties] made their contract."[33] As Alan Schwartz and Bob Scott noted in their canonical article, *Contract Theory and the Limits of Contract Law*, this question in theory has a "correct answer."[34] In practice, however, it is not always easy or possible to know what it is. Lacking a time machine, adjudicators traditionally have stitched together an answer using imperfect evidence—a mix of the contract's text, the parties' statements about the deal (whether from before, during, or after its formation),[35] market data,[36] and some hunches about fairness and efficiency under the circumstances.[37]

---

[30] For defenses of contextualism, see Jeffrey W. Stempel & Erik S. Knutsen, *Rejecting Word Worship: An Integrative Approach to Judicial Construction of Insurance Policies*, 90 U. CIN. L. REV. 561, 600-601 (2021); Jeffrey W. Stempel, *Unmet Expectations: Undue Restriction of the Reasonable Expectations Approach and the Misleading Mythology of Judicial Role*, 5 CONN. INS. L.J. 181, 183-84 (1998).

[31] In some industries, the evidence that parties would prefer that later decisionmakers incorporate context is robust. William Hoffman, *On the Use and Abuse of Custom and Usage in Reinsurance Contracts*, 33 TORT & INS. L.J. 1, 3 (1997) (origin of nonintegrated contracts); William Hoffman, *Facultative Reinsurance Contract Formation, Documentation, and Integration*, 38 TORT TRIAL & INS PRAC. L.J. 763, 836-37 (2003) (explaining why parties prefer custom).

[32] Bruce v. Blalock, 241 S.C. 155, 161, 127 S.E.2d 439, 442 (1962) ("In construing the contract the Court will ascertain the intention of the parties . . . as well as the purposes had in view at the time the contract was made.").

[33] STEVEN J. BURTON, ELEMENTS OF CONTRACT INTERPRETATION § 1.1, at 1.

To put it another way, almost all jurists agree that the goal of contract interpretation—its real ambition—is to be a prediction machine.[38] Nonetheless, interpretation is "the least settled, most contentious area of contemporary contract doctrine and scholarship."[39] That's because of the many problems it seeks to solve. As Greg Klass puts it, jurists ask (1) whose meaning counts, (2) what type of meaning matters (local/majoritarian, semantic/pragmatic), and (3) what facts determine the legally relevant meaning.[40] These questions map, imperfectly, onto distinctions between textualists and contextualists. And, at the bottom of the well, contractual interpretation resolves questions of claims to judicial power, and thus legitimates violence.[41] The result is that parties contesting how to interpret contracts are really arguing about what outcomes are just, not merely which are more likely to lead to parties getting what they want.

---

[34] Alan Schwartz & Robert E. Scott, *Contract Theory and the Limits of Contract Law*, 113 YALE L.J. 541, 568 (2003) ("There is a consensus among courts and commentators that the appropriate goal of contract interpretation is to have the enforcing court find the 'correct answer.'"); Alan Schwartz & Robert E. Scott, *Contract Interpretation Redux*, 119 YALE L.J. 926 (2010). For criticisms, see Adam B. Badawi, *Interpretive Preferences and the Limits of the New Formalism*, 6 BERKELEY BUS. L.J. 1 (2009); Shawn J. Bayern, *Rational Ignorance, Rational Closed-Mindedness, and Modern Economic Formalism in Contract Law*, 97 CAL. L. REV. 943 (2009); Robin Bradley Kar and Margaret Jane Radin, *Pseudo-Contract and Shared Meaning Analysis*, 132 HARV. L. REV. 1135, 1182-92 (2020) (arguing that sophisticated parties would not and do not prefer acontextual readings).

[35] Stephen F. Ross & Daniel Trannen, *The Modern Parol Evidence Rule and its Implications for New Textualist Statutory Interpretation*, 87 GEO. L.J. 195, 196-97 (1995) (noting disagreement between Williston and Corbin on parol evidence).

[36] JOHN BOURDEAU, PAUL M. COLTOFF, JILL GUSTAFSON, GLENDA K. HARNAD, JANICE HOL-BEN, SONJA LARSEN, LUCAS MARTIN, ANNE E. MELLEY, KARL OAKES, KAREN L. SCHULTZ & ERIC C. SURETTE, AMERICAN JURISPRUDENCE § 219 (2nd ed. 2023) ("Under the Uniform Commercial Code, a course of dealing between the parties . . . may give particular meaning to, and supplement or qualify, terms of an agreement.").

[37] Omri Ben-Shahar, David A. Hoffman and Cathy Hwang, *Nonparty Interests in Contract Law*, 171 U. PA. L. REV. 1095, 1017-1129 (2023) (describing courts use of public interests in interpreting contracts).

[38] Schwartz and Scott, *supra* note 34, at 568 (noting "consensus" about the "appropriate goal"). There are exceptions. Eyal Zamir, for example, argues that interpretation should adhere to moral and social norms, partly because they are more likely to reflect the parties' true intent, and partly because only those contracts are worth enforcing. *Cf.* Eyal Zamir*, The Inverted Hierarchy of Contract Interpretation and Supplementation*, 97 COLUM. L. REV. 1710, 1777–88 (1997). Other common reasons to deviate from the parties' intentions include attempts to incent clearer drafting, to share valuable information, and to facilitate standardization. *See, e.g.,* Ian Ayres, *Default Rules for Incomplete Contracts,* in 1 THE NEW PALGRAVE DICTIONARY OF ECONOMICS AND THE LAW 585 (Peter Newman ed., 1998) (reviewing the economic theories for the design of default rules). It is inevitable that the parties at times will choose not to think about a relevant possibility to minimize transaction costs or permit a deal. Therefore, when we say that the goal is prediction, consider it the beginning, rather than the end, of interpretation.

But putting aside normative questions, even basic operational empirics about interpretation—the prediction questions everyone agrees are at the core—are *hard*. Prediction is difficult, and mistakes are inevitable. Accuracy—in the sense of thinking that we really got as close as we could to knowing what the parties would have said—trades off against cost and certainty. Efficiency-minded scholars have repeatedly argued that as the amount of evidence offered to prove the parties' contemporaneous-to-contracting meaning increased, so does expense across several domains.[42]

As a first cut at that cost, consider that when parties are permitted to adduce additional sources of interpretative evidence, they also increase the range of defensible answers from the tribunal. This means that it becomes harder to know what the fact-finder will do—their ability to choose unexpected meanings waxes with the evidentiary inputs.[43] But worse, both parties and fact-finders are motivated in how they offer and process evidence.[44] In a regime that permits more evidence, parties will offer evidence

---

[39] Ronald J. Gilson, Charles F. Sabel & Robert E. Scott, *Text and Context: Contract Interpretation as Contract Design*, 100 CORNELL L. REV. 23, 25 (2014); Schwartz and Scott, Redux, *supra* note 34.

[40] *See* Gregory Klass, *Contracts, Constitutions and Getting the Interpretation-Construction Distinction Right*, 18 GEO. J. L. & PUB. POL'Y 13, 24-28 (2020).

[41] *See* Robert M. Cover, *Violence and the Word*, 95 YALE L.J. 1601, 1601 (1986).

[42] *See generally* Gregory Klass, *Contract Exposition and Formalism*, GEORGETOWN UNIVERSITY LAW CENTER SCHOLARSHIP @ GEORGETOWN LAW 63 (2017) ("The more evidence one allows into interpretation, the less certain the outcome. The costs of such uncertainty in the contractual setting can be especially high."); Schwartz & Scott, *supra* note 34, at 580 (2003) ("Expanding the evidentiary base is not costless, however. The parties, therefore, face a tradeoff between the efficiency of increased accuracy and the inefficiency of increased contract-enforcement costs.").

[43] Klass, *id.* ("A party that wants to organize its behavior . . . needs to be able to predict how an adjudicator will later interpret that agreement. To the extent thicker interpretive rules reduce predictability, they impose an additional cost . . . .").

[44] Christoph Engel, *Judicial Decision-Making. A Survey of the Experimental Evidence*, MPI COLLECTIVE GOODS DISCUSSION PAPER, No. 6. 5, (2022) (noting that even when decision makers are

that favors their view, sometimes unconsciously motivated to avoid presenting data that favors the other side;[45] factfinders, equally subject to motivated cognition, will process new evidence in biased ways.[46]

At the same time, as the types of evidence relevant to contract interpretation become more capacious, parties will seek to introduce more evidence at trial, raising the costs of litigation.[47] These costs may be significant, even in dispute resolution forums like arbitration that are built to resolve cases quickly and cheaply.[48] The interpretation arms race has led scholars to model when parties would prefer to spend money ex ante on more specified text, rather than spend *ex post* on litigation.[49] That is, to pre-commit to methodologies which are less accurate but more efficient.

This is all familiar territory. Now, consider what interpretative methodologies have been on offer to calibrate between predictive accuracy and virtues that center around certainty and efficiency. Like other legal extrapolative enterprises, interpretation has developed two basic methods to solve for the predictive question in the absence of

---

motivated to be impartial, bias has been shown to sneak in inadvertently via race, gender, ideology, and the stereotype that tattoos are typical for criminals.); Lawrence M. Solan, Terri Rosenblatt & Daniel Osherson, *False Consensus Bias in Contract Interpretation,* 108 COLUM. L. REV. 1268, 1269 (2008) (explaining that "false consciousness bias" may cause contracting parties not to recognize different interpretations of their agreement until litigation, at which point judges fall victim to the same bias.).

[45] Schwartz & Scott, *supra* note 34, at 607 (2003) (claiming that under standards allowing for recovery of "commercially reasonable" costs and investments, parties would always claim their costs were higher and their investments reasonable).

[46] Solan, Rosenblatt & Osherson, *supra* note 44, at 108 ("Susceptibility to false consensus bias places judges engaged in the interpretation of contractual language at risk of erroneous decisionmaking.").

[47] For some evidence on this process in the courts, *see* Lisa Bernstein, *Custom in the Courts*, 110 NW. U. L. REV. 63 (2015) (showing that courts accept evidence of custom that isn't systematic even in commercial disputes).

[48] Richard A. Posner, *The Law and Economics of Contract Interpretation*, 83 TEXAS L. REV. 1581, 1605-1606 (2004) (arguing that commercial arbitration, where the arbitrator uses commercial common sense to predict intent rather than asking the parties to present evidence, may be preferable when the written contract does not make the parties' intentions immediately clear because it allows the parties to avoid extra expenses).

[49] Ronald J. Gilson, Charles F. Sabel & Robert E. Scott, *Braiding: The Interaction of Formal and Informal Contracting in Theory, Practice, and Doctrine*, 110 COLUM. L. REV. 1377, 1391 n.35 (2010) ("If conditions are unlikely to change much in the future (the level of uncertainty is low), and thus the ex-ante cost of writing contract rules is low relative to the anticipated gains, the parties' most cost-effective strategy is to write a complex, rule-based contingent contract.").

the ability to travel to the time of contracting.[50] These methods, with the shorthand names textualism and contextualism, in the real world are represented by the courts in New York and California, respectively.[51]

New York's textualist judges focus the contract: they take its words as the canonical source of the parties' meaning and abjure other sources of evidence as predictive grist. Textualists try to use the common sense meaning of words, using dictionaries to obtain the public meaning of the words the parties chose, and grammatical and lexical tools to understand how the words, when collated, create obligation.[52] Textualism has known advantages, including forcing the parties to think carefully about what they mean, and to use contract words in ordinary ways.[53] This ideological approach to contract interpretation resembles that same concept in statutory and constitutional interpretation:[54] though it is less politically valanced, it is equally ascendent.[55]

The linguistic textualist project has long been controversial. To begin with, the method of brute sense plain meaning primes judges to overconfidently believe that their beliefs and conclusions are more common than they in fact are.[56] As Arthur Corbin put it long ago, "when a judge reads the words of a contract he may jump to the instant and confident opinion that they have but one reasonable meaning and that he knows what it is."[57] Empirical work—experimental[58] and sociological[59]—has since found that judges doing plain meaning analysis disagree with each other and with lawyers about things they thought obvious.

---

[50] John F. Manning, *What Divides Textualists from Purposivists?*, 106 COLUM. L. REV. 70, 75 (2006) (arguing that textualism and purposivism remain meaningfully distinct modes of statutory interpretation); *see generally* Eric A. Posner, *The Parol Evidence Rule, the Plain Meaning Rule, and the Principles of Contractual Interpretation*, 146 U. PA. L. REV. 533 (1998) (defending textualist approaches in contract law).

[51] Klass, *supra* note 40, at 29 (distinguishing New York and California archetypes).

[52] Joshua M. Silverstein, *Contract Interpretation Enforcement Costs: An Empirical Study of Textualism Versus Contextualism Conducted Via the West Key Number System*, 47 HOFSTRA L. REV. 1011, 1014 (2019) ("'Textualist' judges and commentators argue that the interpretation of contracts should focus primarily on the language contained within the four corners of written agreements."); Gilson, Sabel & Scott, *supra* note 49, at 40 ("Textualist arguments accordingly focus on the insight that, for legally sophisticated parties who write bespoke contracts, context is endogenous; the parties can embed as much or as little context into a customized agreement as they wish, and they can do so in many different ways."); Uri Benoliel, *The Interpretation of Commercial Contracts: An Empirical Study*, 69 ALA. L. REV. 469, 472-473 (2017) (noting importance of ambiguity).

[53] Schwartz & Scott, *supra* note 34, at 572.

[54] For a discussion of the differences between statutory and contract textualism, see William Baude & Ryan D. Doerfler, *The (Not So) Plain Meaning Rule*, 84 U. CHI. L. REV. 539, 563-65 (2017). For an insightful argument that interest in contract interpretation has waned relative to statutory interpretation, see Karen Petroski, *Does it Matter What We Say About Legal Interpretation?* 43 MCGEORGE L. REV. 359, 382 (2019).

[55] Ethan J. Leib, *The Textual Canons in Contract Cases: A Preliminary Study*, 2022 WIS. L. REV. 1109 (2022) (studying the use of textualist canons in contract interpretation); J. Stempel & Knutsen, *supra* note 30, at 565–66 ("In short, textualism has been resilient and ascendant in the 40 years of the post-Restatement era.").

Critics also charge textualists with incoherence about ambiguity.[60] To reach the safe shoals of plain meaning, textualists ask first if the language is unambiguous.[61] But while textualism provides tools to discover ambiguities, in practice, critics charge, it fails to prioritize one plausible interpretation over the other in practice. It appears to simplify interpretative disputes, but in reality sometimes facilitates expensive, biased battles over extrinsic evidence.[62]

But even outside of ambiguity, textualism's basic methodological tools are remarkably underdeveloped. Scholars often blame the humble dictionary.[63] Courts

---

[56] *See infra* at text accompanying notes 109 through 110.

[57] ARTHUR LINTON CORBIN, CORBIN ON CONTRACTS § 535 (rev. ed. 1960)

[58] Solan, Rosenblatt & Osherson, *supra* note 44, at 1285–94 (finding that we overestimate our sense of whether others will agree about contract interpretation).

[59] John F. Coyle, *The Canons of Construction for Choice-of-Law Clauses,* 92 WASH. L. REV. 631, 682–87 (2017) (showing that in the absence of a systematic survey, judges can interpret contract language in ways that conflict with the parties' intentions).

[60] *See* Lawrence M. Solan*, Pernicious Ambiguity in Contracts and Statutes*, 79 CHI-KENT L. REV. 859, 859 (2004) (describing problems with the concept of ambiguity).

[61] 11 Williston on Contracts § 33:43 (4th ed.) ("When patent ambiguities are found by a court that adheres to the traditional distinctions, they will be resolved by the rules of interpretation or not at all."). Those supposed rules of interpretation reference §30:4, where they turn out to combine extrinsic evidence, contract purpose, and rules of construction.

[62] Ward Farnsworth, Dustin F. Guzior & Anup Malani, *Ambiguity About Ambiguity: An Empirical Inquiry into Legal Interpretation*, 2 J. LEGAL ANALYSIS 257, 271 (2010) (policy preferences drive ambiguity) (statutory); Schwartz & Scott, *supra* note 34, at 570 n.55 ("Courts seldom distinguish between 'vague' and 'ambiguous' terms . . . . More narrowly, however, a word is vague to the extent that it can apply to a wide spectrum of referents, or to referents that cluster around a modal 'best instance,' or to somewhat different referents in different people.").

[63] Thomas R. Lee & Stephen C. Mouritsen, *Judging Ordinary Meaning*, 127, YALE L.J. 788, 801, 810-11 (2018) (identifying several problems with dictionaries, including their failure to define words in terms of "prototypes" and the inconsistency of definitions across dictionaries); Stephen C. Mouritsen, *The Dictionary Is Not a Fortress: Definitional Fallacies and a Corpus-Based Approach to Plain Meaning*, 5 BYU L. REV. 1915, 1919 (2010) (describing "widely shared" false views about

doing textualism are sometimes reversed for failing to use one.[64] But it's an imprecise tool for discerning the parties' intent at the drafting stage. Selecting between dictionaries is a value-laden act,[65] and even within a single volume, dictionaries do not provide a single plain, or majoritarian meaning of words.[66] Critically, dictionary definitions are blind even to internal context, those other parts of the document or statute that textualists *do* embrace.[67] As Kevin Tobia demonstrated, definitions can be poor trackers of actual usage, a point well understood by anyone not adding tomatoes to a fruit salad.[68]

Dictionary-thumping jurists face two opposing critiques: they bind themselves too much[69] and but also too little.[70] The first strips the judicial process of its nuanced nature, the latter breeds gamesmanship and bias.[71] This critique is (to be fair) a little overheated. Sure, judges take dictionaries seriously,[72] but they also freely admit that

---

dictionaries); Lawrence Solan, *When Judges Use Dictionaries*, 68 AM. SPEECH 50, 50 (1993) ("[W]e commonly ignore the fact that someone sat there and wrote the dictionary, and we speak as though there were only one dictionary, whose lexicographer got all the definitions 'right' in some sense that defies analysis."); Samuel A. Thumma & Jeffrey L. Kirchmeier, *The Lexicon Has Become A Fortress: The United States Supreme Court's Use of Dictionaries*, 47 BUFF. L. REV. 227, 276 (1999) ("[A]s with the other steps in the Court's general process of using dictionaries, selecting a specific definition for a term can be problematic, at times appears to lack principled guidance and can determine the out-come of a case.").

[64] Lorillard Tobacco Co. v. Am. Legacy Found., 903 A.2d 728, 738 (Del. 2006) (reversing for failure to follow dictionary).

[65] Lee & Mouritsen, *supra* note 63, at 807 ("A common use of a dictionary involves simple cherry-picking.").

[66] *Id.* at 810-11 ("We cannot tell from the opinion whether the written translator sense of interpreter is less often listed in a real 'survey' of dictionaries because we are not presented with an actual survey of dictionaries.").

[67] 11 WILLISTON ON CONTRACTS § 32:5 (4th ed.) ("A contract will be read as a whole and every part will be read with reference to the whole"); Bradley C. Karkkainen, *"Plain Meaning:" Justice Scalia's Jurisprudence of Strict Statutory Construction,* 17 HARV J. L. & PUB. POL'Y. 401, 407 (1994).

[68] Kevin P. Tobia, *Testing Ordinary Meaning,* 134 HARV. L. REV. 726, 797-99 (2020).

[69] Nicholas S. Zeppos, *Judicial Review of Agency Action: The Problems of Commitment, Non-Contractability and the Proper Incentives*, 44 DUKE L.J. 1133, 1143 (1995) ("fanatical" devolution to dictionaries).

[70] *See* Mouritsen, *supra* note 31, at 1930. (critiquing dictionaries as weak source of plain meaning and for the absence of context); *Jordan v. De George*, 341 U.S. 223, 234 (1951) (Jackson, J., dissenting) (calling dictionaries "the last refuge of the baffled judge").

[71] Lee & Mouritsen, *supra* note 63, at 798 ("The concern here is that even if we could settle on a theory of ordinary or plain meaning, we are unsure how to assess it.").

[72] *See e.g.*, Matter of the Liquidation of Am. Mut. Liab. Ins. Co., 440 Mass. 796, 801, 802 N.E.2d 555 (2004) ("Normally, a dictionary definition of a term is strong evidence of its common

dictionaries are not "infallible."[73] Even Learned Hand cautioned, "it is one of the surest indexes of a mature and developed jurisprudence not to make a fortress out of the dictionary."[74] Dictionaries are normally under-determinative of outcomes, and this is a virtue rather than a vice. As we shall claim, this virtue is equally shared by generative interpretation.

Similarly, the canons of interpretation themselves are difficult to defend empirically.[75] These canons are traditionally known by their evocative Latin names—*in pari materie, expressio unius est exclusio alterius, ejusdem generis, contra proferentem, generalia specialibus non derogant*—and they are used to fill dictionaries' gaps.[76] They try to address the problem of context by giving heuristics to parse the parties' proffered meanings.[77] Popular with judges but absent from the Restatement,[78] scholars criticize them as essentially ad hoc.[79] There is no obvious way to know what to do when different canons lead to different outcomes, meaning that they offer the same kinds of degrees of freedom as dictionaries do.

Nor is it clear that the contractual linguistic canons are rooted in how parties

---

meaning."); *see also* Brigade Leveraged Cap. Structures Fund Ltd. v. PIMCO Income Strategy Fund, 466 Mass. 368, 374, 995 N.E.2d 64, 69 (2013).

[73] Cyprus Plateau Min. Corp. v. Commonwealth Ins. Co., 972 F. Supp. 1379, 1384 (D. Utah 1997) ("Dictionaries, while not infallible (or even consistent), are general guides to common usage.").

[74] Cabell v. Markham, 148 F.2d 737 (2nd Cir. 1945).

[75] Farshad Ghodoosi & Tal Kastner, *Big Data on Contract Interpretation,* U.C. DAVIS L. REV. 1, 58 (forthcoming 2024) (highlighting the issue of precedent around the use of canons being deployed without regard to the context in which the precedent arose); Leib, *supra* note XX, at 1110 ("Few scholars or lawyers believe they are applied consistently enough to be reliable in predicting case outcomes . . . .").

[76] *See generally* Edwin Patterson, *The Interpretation and Construction of Contracts,* 64 COLUM. L. REV. 833, 852-55 (1964) (identifying canons of contract interpretation).

[77] The canons of contract interpretation are to be distinguished from the canons of construction in statutory interpretation. As Ryan Doerfler has explored, those canons have been subject to a rehabilitative project over the last generation. Ryan D. Doerfler, *Late-Stage Textualism*, 2022 SUP. CT. REV. 267, 269 (2022).

[78] Ethan J. Leib, *The Textual Canons in Contract Cases: A Preliminary Study*, 2022 WIS. L. REV. 1109, 1112 (2022) ("Yet the Restatement does not treat the textual canons like *expressio unius, ejusdem generis,* or *noscitur a sociis* at all"); Ghodoosi & Kastner, *supra* note 75, at 48 ("While substantive canons have remained roughly in equilibrium over time, the chart below demonstrates a trend in which the invocation of textual canons by courts across contract cases is increasing.").

[79] Karl N. Llewellyn, *Remarks on the Theory of Appellate Decision and the Rules or Canons about How Statutes Are to Be Construed*, 3 VAND. L. REV. 395, 401 (1950) ("there are two opposing canons on almost every point").

think or write.[80] The extant empirical work on linguistic canons in statutory interpretation suggests that the answer is: they might be, but only some of the time.[81] Now, to be sure, some of the canons, like *contra proferentem*, aren't intended to replicate how the parties would have understood the contract at drafting (if that has meaning in contracts deployed to millions of adherents). These normative canons may, or may not, relate to the parties' contemporaneous intentions.[82] But other canons are intended to reflect ordinary uses of language, and yet have been subject to remarkably little controlled scrutiny.[83]

Notwithstanding its methodological shortcomings, contract textualism is ever more popular.[84] That's so for a whole host of reasons, but none more so than the weakness of its main conceptual rival: contextualism. This familiar alternative starts with the same perspective as textualism: what would the parties have said they meant had we asked them at contract? But contextualism invites parties to offer extrinsic evidence to build depth into the predictive analysis. By doing so, contextualism seeks to privilege accuracy – the parties' *real* intent.

---

[80] Gregory Klass, *Interpretation and Construction in Contract Law* 48, at https://papers.ssrn.com/sol3/papers.cfm?abstract_id=2913228 (2018) ("The communicative content of juristic acts includes the intent to effect a legal change, and therefore to satisfy an altering rule. The acts that generate and alter contractual obligations, in distinction, need not be juristic acts—though they sometimes are. Rules of construction are only sometimes pragmatically prior to contract interpretation, but not always and not pervasively.").

[81] Kevin Tobia, Brian Slocum, & Victoria Nourse, *Statutory Interpretation from the Outside*, 122 COLUM. L. REV. 213, 241-243, 262 (2022) (finding that some linguistic canons are stated overbroadly or inaccurately but many canons do reflect the intuitive judgment of ordinary people); Kevin Tobia & Brian G. Slocum, *The Linguistic and Substantive Canons* 23, at https://papers.ssrn.com/sol3/papers.cfm?abstract_id=4186956 23 (2022) ("providing evidence that some interpretive canons that are traditionally motivated by normative values also have a basis in language").

[82] Christopher J. Walker, *Legislating in the Shadows*, 165 U. PA. L. REV. 1377, 1404 (2017) (arguing that "contra proferentem" is not a method by which the true intent of the parties is determined, but rather, is a decision to impose the burden of ambiguity on the drafter).

[83] Ross & Tranen, *supra* note 6, at 226 ("Descriptive canons are based on the way ordinary people express themselves in English.").

[84] Ghodoosi & Kastner, *supra* note 75, at 49 ("our study provides evidence that textualism is on the rise in contract interpretation."); Aaron D. Goldstein, *The Public Meaning Rule: Reconciling Meaning, Intent, and Contract Interpretation*, 53 SANTA CLARA L. REV. 73, 77 (2013) (arguing that courts have increasingly moved away from the use of extrinsic evidence to help them understand the parties' intent, leaning instead on "objective" manifestations of intent); Mark L. Movsesian, *Formalism in American Contract Law: Classical and Contemporary*, 12 IUS GENTIUM 115 (2006) ("It is a truth universally acknowledged, that we live in a formalist era. At least when it comes to American contract law.").

This approach to interpretation (really, capacious in the types of evidence considered relevant), found its heyday in the 1960s in California and has never been as popular since.[85] The problem with the approach, according to its critics, is that it does not permit the parties to know what meaning a court will assign to the words they write, since the other side can always offer self-serving meanings ex post and, if believable enough, write a new bargain in court to replace the one drafted in the past.[86] Even contextualism's origin story is one of a party suddenly remembering that they actually meant to make the purchase option available only to family members, creditors be damned.[87] Contextualism makes it difficult to lock down meaning ex ante, through merger clauses and the like, which are always subject to later testimonial refutation. Contextualism's consumer protection allure is understandable.[88] But even if contextualism could offer more accuracy, at what cost?[89]

Indeed, scholars often defend textualism on efficiency grounds.[90] Though it may be unclear what parties want interpretative rules to be, it's almost certainly the case that lawyer-drafters prefer textualist to contextualist modes of decision. Eric Posner captures the idea well: parties will often include an explicit merger clause, but few ever bother with an "anti-merger clause."[91] Thus, from the perspective of the litigated cases—those

---

[85] *See* Pac. Gas & Elec. Co. v. G. W. Thomas Drayage & Rigging Co., 69 Cal. 2d 33, 442 P.2d 641 (1968); *see also* Masterson v. Sine, 68 Cal. 2d 222, 436 P.2d 561 (1968).

[86] *Masterson*, 68 Cal. 2d at 231 (Burke, J., dissenting).

[87] *Id.*

[88] Olah v. Ganley Chevrolet, Inc., 2010-Ohio-5485, ¶ 15, 191 Ohio App. 3d 456, 460, 946 N.E.2d 771, 774 (holding that buyers of a vehicle are barred from presenting evidence that the car was represented by the dealer as new because the contract says the vehicle is used).

[89] An admittedly limited survey of enforcement costs did not find meaningful differences between textualist approaches and contextualists ones. *See* Silverstein, *supra* note 52. For an argument that textualism produces higher enforcement costs because of the judge-by-judge variation in outcomes produces more litigation, *see* 6 PETER LINZER, CORBIN ON CONTRACTS § 25.14[B] at 163 (Joseph M. Perillo ed., rev. ed. 2010).

[90] Schwartz & Scott, Redux, *supra* note 34, at 928 & n.3 (2010) ("A strong majority of U.S. courts continue to follow the traditional, 'formalist' approach to contract interpretation"). *But see* Joshua M. Silverstein, *Contract Interpretation and the Parol Evidence Rule: Toward Conceptual Clarification*, 24 CHAP. L. REV. 89, 92 (2020) (arguing that the matter is indeterminate); Silverstein, *supra* note 52, at 1020 ("contracts scholars can also generally be split into textualist and contextualist camps, with a clear majority falling into the latter group"). There is recent evidence that contract scholars prefer contextualism. Eric Martinez and Kevin Tobia, *What Do Law Professors Believe About Law and the Legal Academy*, 112 GEO. L. REV. ___, 42 (forthcoming 2023).

[91] Eric A. Posner, *The Parol Evidence Rule, the Plain Meaning Rule, and the Principles of Contractual Interpretation*, 146 U. PA. L. REV. 533, 571 (1998). As Larry Solan later pointed out, merger clause

between rich and lawyered parties—contextualism is simply harder to defend.

And yet, from a certain perspective, contextualism seems well-positioned for a revival. Recall that even contextualism's critics agree about first-order goal: to figure out what the parties would have meant at contracting. The problems with contextualism are largely centered around motivated testimony and cost, which persuades the fact-finder to ignore the text. But consider: we increasingly live in a world where our thoughts are recorded contemporaneously, whether sent by text, posted on social media, or recorded on TikTok. Such recorded, immutable utterances are cheap to reproduce and appear to courts to be excellent sources of contractual meaning.[92] Defenders of textualism may argue that permitting their use creates uncertainty, but some of the best arguments against contextualism—that it can be abused *ex post*—are weaker than they used to be.[93] And yet, we lack a method to know which excited utterances to privilege, and we should worry that courts' motivated reading will cause them to come to inaccurate or biased understandings.

The debate between textualism and contextualism is old, and scholars have offered various theoretical lenses by which one or the other approach ought to prevail.[94] Most arguments for or against extrinsic evidence turn on hypotheses about what parties would have wanted (had we asked them) and which methods promote social welfare. These arguments are often theoretically rich but empirically poor.[95]

More recently, scholars have offered two new methods, both advancing the certainty values of textualism with a dash of the accuracy interests of contextualism. One

---

analogs in statutory interpretation "are not easy to find." LAWRENCE M. SOLAN, THE LANGUAGE OF STATUTES: LAWS AND THEIR INTERPRETATION 187 (Chicago 2010).

[92] *See* BrewFab, LLC v. 3 Delta, Inc., No. 22-11003, 2022 WL 7214223, at *1 (11th Cir. Oct. 13, 2022) (affirming that a party's text message was a personal guaranty that satisfied Florida's statute of frauds); *see also* Cloud Corp. v. Hasbro, Inc. 314 F.3d 289, 295 (7th Cir. 2002) (finding that a party's e-mails satisfied the UCC's statute of frauds and using these as evidence in support of the claim that the contract had been modified); *see also* Cosby v. Am. Media, Inc., 197 F. Supp. 3d 735, 744 (E.D. Pa. 2016) (holding that tweets may form the basis of a breach of contract claim).

[93] *Cf.* Shawn Bayern, *Contract Meta-Interpretation*, 49 U.C. DAVIS. L. REV. 1097, 1136 (2016) (pointing out that because text messages are informal, they don't satisfy some of the deliberation-inducing virtues that textualists would otherwise place in written documents).

[94] *See* Ross & Tranen, *supra* note 5, at 196-197; *see also* Joshua M. Silverstein, *The Contract Interpretation Policy Debate: A Primer*, 26 STAN. J.L. BUS. & FIN. 222 (2021); *see also* Mark L. Movsesian, *Severability in Statutes and Contracts*, 30 GA. L. REV. 41, 70 n.184 (1995) (noting that the popularity of the major interpretive approaches "ebbs and flows").

[95] Silverstein, *supra* note 52, at 1014 ("The textualist/contextualist controversy cannot be resolved in the abstract. . . . Unfortunately, empirical evidence bearing on this debate is sorely lacking.").

school focuses on the use of corpora of words to predict the meaning of phrases in contractual texts—so-called *corpus linguistics*.[96] To take the prototypical example, consider the following phrase taken from an insurance contract:

> [T]his insurance does not apply to 'bodily injury' [including death] to any person while practicing for or participating in any sports or athletic contest or exhibition that you sponsor."

An insured dies while snorkeling: is that a "sports or athletic contest"? As Stephen Mouritsen observes, the question is not easily answerable using the classic dictionary-and-canon based tools of textualism. And, considering that insurance contracts are drafted by powerful firms, who subject them to regulatory scrutiny, the idea of using extrinsic expressions by either firms or the insured seems hopeless.[97] Instead, Mouritsen suggests that courts could (helped by adversarial presentation by parties) query language databases to establish whether sports and snorkeling appear relatively close to each other in some number of previous examples. That is, to derive the meaning of the word from its common use in previous texts. (The answer is, more or less, that sports are rule-based competitions, while snorkeling is swimming wearing a goofy mask.)[98]

Corpus linguistics is an advance over traditional textualism or contextualism. It provides a methodology that theoretically allows courts to adhere to an objective set of responses when determining the ordinary meaning of words based on their actual usage. Essentially, it's a form of textualism that doesn't rely on dictionary definitions or a battery of canons. It mirrors not the static decisions of lexicographers in their secluded, book-filled offices, but rather is rooted in the public use of words—democratized textualism.

But corpus linguistics is inattentive to context.[99] It can only really compare brief snippets of text, rather than whole documents. Thus, although the method has been repeatedly used in statutory interpretation cases—where the stakes are high, parties are commonly engaged in interpretative battles over short phrases—only one contracts opinion to date has applied the method to date.[100]

---

[96] *See generally* Mouritsen, *supra* note 19, at 1360-1407 (making case).

[97] Christopher C. French, *Insurance Policies: The Grandparents of Contractual Black Holes*, 67 DUKE L.J. ONLINE 40 (2017) (discussing the difficulty of interpreting insurance contracts for evidence of real meaning).

[98] Mouritsen, *supra* note 19, at 1371-74 (CL approach to snorkeling).

[99] *See* Choi, *supra* note 16, at 8, 16-17 (arguing that the context "undermines the core claim of corpus linguistics").

A different constraining approach, advanced by Omri Ben Shahar and Lior Strahilevitz, encourages courts to use survey evidence to decide on the public meaning of certain contractual texts.[101] As they point out, this survey evidence is a second best to the predictive exercise we described above:

> Contracts should have the meaning that the parties to the transaction assign to the text. [But] it is pointless to ask the actual parties in the litigation what the text meant to them when they formed the contract, because they will bend their answers to fit their litigation goals. So the law should instead ask disinterested people just like them.[102]

The authors defend this interesting proposal against various charges.[103] Their core survey case is consumer contracts designed for mass audiences.[104] There, the survey audience and the original adherents are the same (although separated by time), and we should have fewer worries about the parties intending idiosyncratic meanings.[105] But outside of that frame, a problem with the survey approach is that for most litigated contract cases—i.e., commercial cases—the relevant survey audience will be difficult to find, as sophisticated adherents don't take surveys, or will game them, producing the same problems encumbering contextualism.[106]

---

[100] *See* Fulkerson v. Unum Life Insurance Co. of America, 36 F.4th 678 (6th Cir. 2022); *see also* Richards v. Cox, 450 P.3d 1074, 1085-86 (Utah 2019) (Lee, J., concurring) (concurring in majority opinion "to the extent it relies on corpus linguistic analysis" to support constitutional and statutory interpretation). *Cf.* Wilson v. Safelite Group, Inc. 930 F.3d 429, 439 (6th Cir. 2019) (arguing for use of CL in statutory analysis); Caesars Entm't Corp. v. Int'l Union of Operating Eng'rs Local 68 Pension Fund, 932 F.3d 91, 95 n.1 (3d Cir. 2019) (using corpus linguistics to interpret "previously").

[101] Ben-Shahar & Strahilevitz, *supra* note 18; Ian Ayres & Alan Schwartz, *The No-Reading Problem in Consumer Contract Law*, 66 STAN. L. REV. 545 (2014) (advocating empirical testing to identify surprising and problematic provisions in standard form contracts, against which consumers ought to be warned); Ariel Porat & Lior Jacob Strahilevitz, *Personalizing Default Rules and Disclosure with Big Data*, 112 MICH. L. REV. 1417, 1419–20 (2014) (advocating the use of surveys to identify the majoritarian preferences for the design of granular default rules).

[102] Ben-Shahar & Strahilevitz, *supra* note 18, at 1802.

[103] *Id*. at 1802-1813 (making the case).

[104] *Id*. at 1758 (noting focus on consumer contracts).

[105] *Id*. at 1776-1777 (articulating the basic consumer contracts case).

[106] *Cf.* Roberts v. Farmers Ins. Co., 201 F.3d 448 (10th Cir. 1999) ("[W]hat the public expects from an insurance policy is simply not relevant to the legal question of whether the contract is ambiguous.").

Survey evidence is also an expensive adjudicatory technology. Surveys themselves are difficult to conduct: judges would need to rely on their adversarial presentation in the ordinary case. And they are increasingly unreliable: recent work has found that almost a third of online survey respondents use LLMs to complete answers.[107] Surveys based on more collated samples face the same sorts of problems that have bedeviled modern polling: nonresponse bias among parts of the population, difficulties of generalization, and inaccuracy. And even here, attention is scarce. It is hard to survey consumers on a twenty-page policy or to expect anyone filling out a survey for a $5 gift card to attentively consider interdependencies within the contract.

Consequently, though survey methodology is an established technique in trademark cases and could very well be of enormous help in making sense of the meaning of certain consumer contracts, it is unlikely to be a transformative technology in the ordinary contract interpretation case. We are unaware of any cases to date that permit the use of survey evidence to determine contractual meaning.

<div style="text-align:center">*       *       *</div>

In summary, notwithstanding broad agreement about the predictive goal of interpretation, there's also a shared sense that there's something amiss in how jurists balance accuracy and efficiency. Textualism promises the latter, but in practice it often merely supercharges the judge's own overconfident priors. Contextualism promises the former, but probably doesn't deliver it, while eroding parties' ability to plan for court outcomes and making litigation expensive for all but the wealthiest parties. The two most sophisticated modern improvements on these old technologies—statistical plain meaning and survey evidence—promise to rescue textualism from some of its sins, but haven't been taken up in live cases.

Enter large language models.

## II. GENERATIVE INTERPRETATION

The doctrine of reasonable expectations plays a significant role in the regulation of insurance contracts.[108] Generally, the insured reasonable expectations trump the insurance contract's terms. Notoriously, these sorts of cases motivate armchair specula­tion by judges—whose life experience, education, sophistication, and hard-earned cyni­cism systematically diverge from most lay people. Worse, the interpretations we give words appear very certain in our own minds. It is a prime subject to a phenomenon psy­chologists call "false consensus bias."[109] To illustrate the effect, scholars presented con­tract interpretation questions to both laypeople and judges. After asking for their opin­ion, subjects were asked to estimate how many other participants would agree with them. This clever design allowed us to see the actual distribution of answers and how people expected the distribution to look. The results were striking: both laypeople and judges overestimated how common their chosen interpretations were. Judges even overesti­mated how much other judges would agree with them.[110]

---

[107] Veniamin Veselovsky, Manoel Horta Ribeiro, Robert West, *Artificial Artificial Artificial Intelligence: Crowd Workers Widely Use Large Language Models for Text Production Tasks,* ARXIV:2306.07899 (2023) (33-46% of mTurk survey workers use LLMs to complete tasks).

[108] *See generally* Jeffrey W. Stempel, *Unmet Expectations: Undue Restriction of the Reasonable Expectations Approach and the Misleading Mythology of Judicial Role,* 5 CONN. INS. L.J. 181 (1998).

Thus, one of the risks of introspective interpretation is that its products are very sticky and hard to dislodge. This leads to dissent and reversal, and of course, interpreta­tion that defies parties' reasonable expectations. Uncertainty about common interpreta­tion is an appealing case for the use of surveys.[111] And surveys would be of great interpre­tative use, were it not for the practical difficulties which we've just discussed.

Consider *C & J Fertilizer v. Allied Mutual*.[112] The president of C&J, a fertilizer firm, purchased a burglary insurance policy from Allied Mutual. The discussions preced­ing the purchase made it clear that it would not cover an "inside job." The insurance firm insisted that to bring a claim, C&J would have to present hard evidence that a theft was made by a stranger.[113] That idea was embodied in the following promise in the insurance contract:

> [Allied will pay for] the felonious abstraction of insured property (1) from within the premises by a person making felonious entry therein by actual force and violence, of which force and violence there are visible marks made by tools, explosives, electricity or chemicals . . .[114]

---

[109] Joachim Krueger & Russell W. Clement, *The Truly False Consensus Effect: An Ineradicable and Egocentric Bias in Social Perception*, 67 J. PERSONALITY & SOC. PSYCHOL. 596, 596-97 (1994); Brian Mullen, Jennifer Atkins, Debbie S. Champion, Cecelia Edwards, Dana Hardy, John E. Story, and Mary Vanderklok, *The False Consensus Effect: A Meta-Analysis of 115 Hypothesis Tests,* 3 J. EXP. SOC. PSYCH. 262 (1985).

[110] Lawrence Solan, Terri Rosenblatt & Daniel Osherson, *False Consensus Bias in Contract Interpre­tation*, 108 COLUM. L. REV. 1268, 1291 (2008).

[111] *See generally* Stempel, *supra* note 30.

[112] C & J Fertilizer, Inc. v. Allied Mut. Ins. Co., 227 N.W.2d 169, 176 (Iowa 1975).

[113] 227 N.W.2d. at 172.

[114] 227 N.Y.2d at 171.

As it turns out, a burglar robbed the fertilizer plant with style. While leaving some tread marks in the mud, he avoided leaving any other visible signs before absconding with $50,000 worth of fertilizer. The insurance company, denying the claim, argued that by its plain language, the absence of visible marks made by tools (as opposed to tires) meant that it didn't have to pay.

The Iowa Supreme Court, in a contacts casebook staple, held that the exclusion applied in this way violated the insureds' reasonable expectations. No one could have reasonably expected that burglary would be limited only to those leaving visible entry marks.[115] In reaching that view, the Court relied on its own common sense with no empirical grounding. Was it right?

That question triggers the simplest use cases of LLMs as part of the interpretative process. The judge can simply *ask* the model for its assessment. Fantastical only three years ago, today you might be merely whelmed by the model's ability to respond coherently and plausibly to this query. Here's the model's response, edited for readability:[116]

A given business insurance policy covers third-party burglaries but does not extend its coverage to inside jobs. With this in mind, please state your prediction—with the associated level of confidence in parentheses—on the likely expectations of most policyholders under these terms for the following propositions:

---

[115] 227 N.Y.2d. at 177 ("But there was nothing relating to the negotiations with defendant's agent which would have led plaintiff to reasonably anticipate defendant would bury within the definition of 'burglary' another exclusion denying coverage when, no matter how extensive the proof of a third-party burglary, no marks were left on the exterior of the premises. This escape clause, here triggered by the burglar's talent . . . was never read to or by plaintiff's personnel, nor was the substance explained by defendant's agent.").

[116] Code repository here: https://chat.openai.com/share/b068ecd4-2e14-4f91-a500-e4f4509dad38 (last visited July 29, 2023).

| Queries and GPT-4 Responses | |
|---|---|
| Query | GPT-4 Response |
| The policy will provide compensation for losses resulting from a substantiated third-party burglary. | Highly likely (90%) |
| The policy will clearly delineate what is considered an inside job as opposed to a third-party intrusion. | Moderately likely (70%) |
| Even in instances where a third-party burglary can be definitively established, the policy will necessitate physical evidence of forced entry. | Likely (80%) |

*Table 1: GPT-4's estimates of propositions regarding the likely content of the gap in the policy.*

In other words, the model disagreed here with the court's majority opinion. It (like a minority opinion) predicts that policyholders would have expected to have been required to provide some evidence of forceful entry to prove that their job was really not an inside one.

To us these findings are facially plausible: they validate that this cheap and convenient tool could be potentially of use in real cases. But just because the probabilities are reasonable doesn't mean they are accurate. Your intuition should be: prove it! You would want to know more, both about what the model is doing when it produces percentages, and how that methodology fits courts' purposes in interpreting insurance contracts. Let's start there, in Part A. We'll then try some more complicated examples in the remainder of this Section.

### A.        A Gentle Introduction to Large Language Models

When Chat GPT-4 told us that it was 90% likely that the policy would pay in response to a "substantiated third-party burglary," what happened behind the curtain? We're going to give an explanation a shot here, knowing that doing so is difficult in part because LLM technology is complex and rapidly changing. Essentially, LLMs create a statistical model of how words connect by training on torrents of existing texts, some historic and some artificially derived.[117]

---

[117] Synthetic data is growing in importance, and sometimes may improve model quality. John Jumper et al., *Highly accurate Protein Structure Prediction With AlphaFold*, NATURE 583, 587-89 (2021) (noting how training the data using synthetic data improved the model's accuracy significantly).

In the common case, LLMs take user input in the form of text and produce an output, also in the form of text. Behind the scenes, the model takes the text and transforms it into numbers. This is essential, because (superficially) computers cannot read text. Because numbers can encode more information than letters, they are more valuable in that they allow computers to perform mathematical operations. This is easy to see in the case of ambiguities: duck is both a noun and a verb. But in a number system, we can use prefixes like 20 for verbs and 10 for nouns, so we can encode the word duck twice. One is, say, 201 and the other 101, to designate the disparate meanings and disambiguate it.[118]

This simple illustration understates the promise of embeddings.[119] Rather than assigning a single number to each word, machine learning models transform them into strings of number-pairs—each one signaling some piece of meaning. The length of such vectors is very long; one of the latest models in common use employs a vector with 12,288 number-pairs.[120] For simplicity of exposition, suppose you had a list of common animals and had a two-dimensional vector to describe them. One dimension could be number of feet; another could be if they lived on land or sea. This would produce vectors that we can visualize below:[121]

---

[118] This, in a sense, is what standard English dictionaries do, at least if one were to number the words by order of appearance.

[119] For a description of embeddings (although without the attention mechanism) see Choi, supra note 16, at 20-22.

[120] Nils Reimers, *OpenAI GPT-3 Text Embeddings – Really a New State-of-the-Art in Dense Text Embeddings?*, MEDIUM, Jan. 28, 2022.

[121] Sea turtles have flippers, not legs. In a more sophisticated representation, we might have adopted a more continuous representation of feet, where flippers are closer to feet than they are to, say, tails.

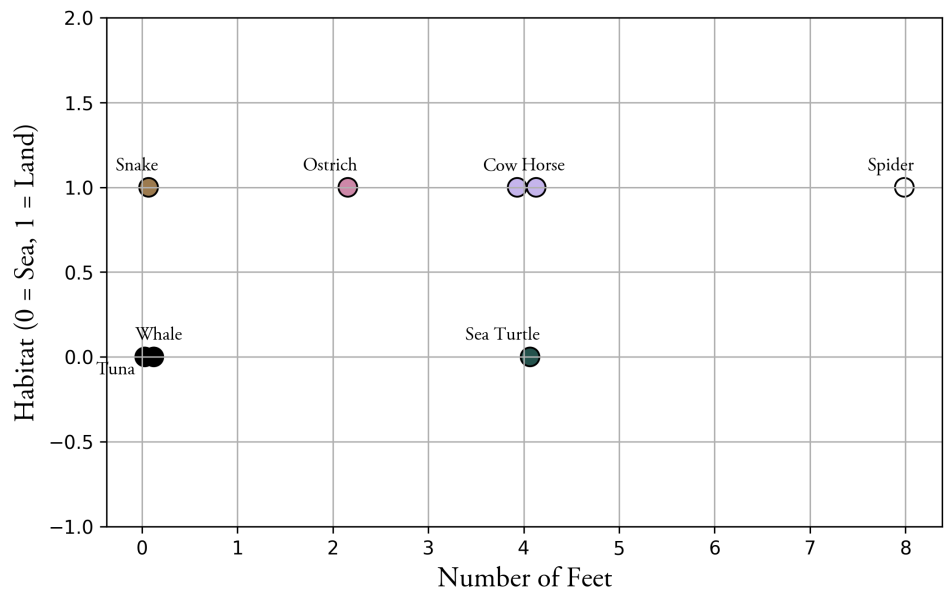

*Figure 2: An illustration of the value of encoding meaning via simple embeddings*

What makes vectors so powerful is that they allow us to capture not only semantics, but also a syntactic relationship to other words. Horses and cows, in our very simplistic schema, are closer to each other than they are to whales or tuna fish. The snake, always awkward, occupies its own category. If we were to add lizards, we would spot the emergence of a distinct category of reptiles, alongside the land mammals. Now, suppose you did the same with over 10,000 dimensions. You can imagine the insights that might result when words are described along such complex dimensions.

Making words dimensional has proved powerful in many machine learning tasks, but was insufficient to power the new LLM revolution. What was needed was the idea of attention.[122] Read the following sentences:

> "Shohei Ohtani felt the stress. In a desperate attempt, he swung the bat."

You intuitively grasp that they mean that Ohtani lifted his bat and used it to swing at the baseball. But how do you know that this was right, and not that Ohtani had swung at a mammal? As Amelia Bedilia taught us, it's possible to turn many normal phrases into misadventures if you ignore context. We know that *swung* typically is associated with objects, not animals. And we connect *bat* with Ohtani, a baseball player, which further solidifies our interpretation of the sentence as referring to the object. In other words, our minds naturally pay attention to the context of the word to infer the meaning of any specific word.

---

[122] This idea was most powerfully described in a 2017 paper. Ashish Vaswani, Noam Shazeer, Niki Parmar, Jakob Uszkoreit, Lloion Jones, Aidan N. Gomez, Lukasz Kaiser, and Illia Polosukhin., *Attention is All You Need*, ARXIV: 1706.03762 (2017).

An LLM's *transformers* seek to achieve the same thing with respect to vectors.[123] The attention mechanism assigns a number (via positional encoding) and an initial vector to each word in a sentence. Then it assesses which words—say bat or swung—shed light on its meaning.[124] In the sentence above, words like stress and felt are not particularly relevant; but both swung and Shohei Ohtani matter. This allows the model to assign an attention score to each component in the input (relative to the word under analysis) and then reweigh the encoding of a word relative to the words that are relevant to its interpretation. This means that words do not have stable embedding (as in the older models), but rather, the embedding changes based on the specific context in which they are presented.

These ideas are combined to train a model. A model refers to a collection of parameters (mostly ones called "weights" and "biases") organized in a specific way whose values are used to transform the input into the model's output. Modern language models contain hundreds of billions of such parameters, hence their common designation as "Large Language Models."

Language models are trained with some objective function, a task which they try to achieve and on which they are evaluated. In the context of most LLM models, the goal is prediction. The model is presented with the sentence "Shohei Ohtani felt the stress. In a desperate attempt he swung the [?]" and then the model predicts which word would come next. If the model is not calibrated, it might guess *lamp* or *materiality*. These are not likely correct answers, and the model will be fed back the result of an optimization technique called gradient descent which will instruct the model on how to update its parameters to make a better guess next time. This process repeats itself until the model learns that *bat* follows with 70.14% probability, *base* with 25.13%, *axe* with 0.53%, *club* with 0.51%, and so on.[125]

---

[123] Sebastian Raschka, Yuxi (Hayden) Liu, and Vahid Mirjalili. Machine Learning with PyTorch and Scikit-Learn (2022).
[124] This is a simplification. Words are assigned a tensor that includes their embedding, positional encoding, and key, value and query values (which are necessary for the attention mechanism).
[125] Based on actual predictions of the Tex-Davinci-003 model with temperature 0.7, max length = 256, top p = 1, 0 frequency or presence penalty and best of 1.

What does *learning* mean here? The answer is both simple and complex. The simple answer is that after training, the model has developed a high complexity probability matrix, comprised of billions of parameters arranged in matrices. It conducts various (fairly simple) algebraic operations to create from a sentence like "Hello, how are you__" a prediction that the next highest probability word would be "doing." The complexity comes from the fact that these parameters are effectively encoded in large, inscrutable matrices whose meaning is wickedly hard to decipher, and whose organization is alien. They do not explain the *why* of their predictions.

You may have read, but would be wrong to conclude, that because the goal is to assign probability to the next word, these models simply imitate text they have seen elsewhere or only develop a superficial model of the world. These models exhibit originality—sometimes making up facts, while other times developing entirely new but responsive text. And yet, we should be very hesitant to call the model's function *understanding*, even if to a first approximation this is a useful metaphor. Recent work on very simple transformers has shown that memorization is not a viable technique for complex models.[126] As large as they are, the models are much smaller than the data they are trained on. And so, models necessarily seek deeper representation of the information they train on. This is not unlike how humans read books, learn from them, but cannot recite them.

Finally, consumer-facing chatbots simply invite the user to chat with the model directly. Behind the scenes, however, the model's behavior is calibrated by settings called "hyperparameters."[127] The details are quite technical, but one of those hyperparameters is of specific interest. The model has "temperature" settings that can be adjusted from low to high. The lower the model's temperature, the more predictable its output.[128] A very low temperature ensures that the model always outputs the same answer to the same query. A higher one introduces more randomness and outputs that you might think of as "creative."

So far, so good. Now let's return to our question: what is the model doing when it assigns a 90% probability to the likelihood that a reasonable person would expect an insurance payment? The first step for the model is to convert the query we entered to numbers (really, tensors). The next step is crucial: now the model *attends* to the context of words and uses it to adjust their meaning. If the model sees the word "premium" in the current context, it will know to adjust its meaning away from dictionary meanings such as "high quality" and towards "consideration for an insurance policy."[129]

---

126 *See* Alethea Power, Yuri Burda, Harri Edwards, Igor Babuschkin, Vedant Misra, Grokking: Generalization Beyond Overfitting on Small Algorithmic Datasets, ARXIV: 2201.02177 (2023)

127 The term "hyperparameter" is necessary to distinguish the model's own parameters, from the parameters that define its training and operation.

128 For a friendly technical review, see FRANÇOIS CHOLLET, DEEP LEARNING WITH PYTHON, CHAPTER 12.1 (Second Edition, 2021).

Armed with a contextual understanding of the query, the model can now run through its vast internal network of parameters and calculate what is the most likely word (really, token) that would follow next. It will assign infinitesimally low probabilities to words that relate to gardening or makeup, but will assign increasingly higher probabilities to words that relate to the insurance context. Once the model determines the most likely continuing words, it orders them by relevancy. In low temperature settings, the model will always select the word with the highest probability to follow, but as we increase the temperature it will occasionally pick other words as well. When the model chooses 90% it is because it predicts that this number is the most likely continuation of the text preceding it.

This explanation skips over the hardest question, which is *why* the model assigns the highest probability to 90%. The honest answer is quite unsatisfactory: it picked this number because based on its vast training data and internal statistical model, it found that 90% is a more likely continuation than 10%. This is nothing like an explanation a human would give, where reasons and factual considerations would be provided. The model's outputs are a brute statistical fact. It is possible to ask the model to justify itself. And the model will diligently reply with an answer. But it is critical to understand that whatever the model tells you, it is really no explanation at all. It is a *prediction* of what explanation is likely to follow the query. So, working with LLMs admittedly requires a leap of faith, a realization that no better explanation is forthcoming than long inscrutable matrices that produce predictions.

B.                    *LLMs as a Source of Contractual Meaning*

With a grasp of the technology in hand, let's work through some more quotidian examples of its potential use outside of the insurance context. Textualists—as we've described—think that texts have an inherent plain meaning, at least within the context of the written document. The problem is deciding what it is, and whether our intuitions are representative. LLMs may serve as powerful tools to uncover those answers.

---

[129] *Premium*, MERRIAM-WEBSTER DICTIONARY, (July 27, 2023), https://www.merriam-webster.com/dictionary/premium.

We'll start with the divorce of Jennie and Mark Famiglio. Jennie and Mark entered into a prenup before getting married, which committed to a sliding scale of payments from Mark to Jennie if they divorced, tied to the length of their union. Section 5.3a read:

> 5.3. JENNIE's Benefits and Obligations. If the marriage ends by dissolution of marriage or an action for dissolution of marriage is pending at the time of MARK's death, then JENNIE shall receive the additional benefits and obligations described in 5.3.a. through d.
>
> > a. MARK shall pay to JENNIE, within ninety (90) days of the date either party files *a Petition for Dissolution of Marriage* the amount listed below next to the number of full years they have been married *at the time a Petition for Dissolution of Marriage is filed.*[130]

Although Jennie filed for divorce after seven years, she never served the petition and later voluntarily dismissed the action. After ten years, she filed again, and meant it. Under the prenup, seven years of marriage entitled her to $2.7 million; ten years a whopping $4.2 million. The parties were left with a consequential but basic interpretive question: when the prenup mentions the number of years at the time "*a*" petition is filed—did the parties mean the *first* petition or the *ultimate* one?

Neither party thought witnesses were necessary, as both understood *a Petition* to be unambiguous (and favoring their side). Unfortunately for Jennie, a Florida appellate court ruled against her.[131] Relying in part on dictionaries, it emphasized that "a" is an indefinite article. Ordinarily, the court stated, when people predicate a condition on an indefinite event, they mean its first occurrence. Thus, imagine if a golf course posts a rule: "when a thunderstorm approaches, you must end your golf game."[132] That would be "universally understood ... to mean the first time a thunderstorm approaches." And so, "a" petition filing simply must mean the first one filed. The court's method of proof seems sensible. But is it right?

We presented GPT with the prenuptial agreement and asked it: If one of the parties files a divorce petition, withdraws it, and then a few years later a new petition is filed, what date determines the number of full years of marriage: the first filing or the second one? It produced a sentence that essentially supported Jeannie's view. But to illustrate how the model can help courts be more precise, we can freeze the output in time and take a peek under the hood, as Figure 3 illustrates.

---

[130] 279 So.3d. at 737.
[131] *Id.* at 742
[132] *Id.* at 743.

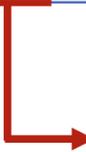

The second filing would determine the number of full years of marriage

| Word | Probability |
| --- | --- |
| second | 94.72% |
| date | 4.44% |
| first | 0.68% |
| number | 0.13% |
| amount | 0.01% |

*Figure 3: Davinci-003, temp=1, top-p=1, frequency and repetition penalty =0, best of 1, full spectrum, presented with Famiglio facts and asked "If one of the parties files a divorce petition, withdraws it, and then a few years later a new petition is filed, what date determines the number of full years of marriage: the first filing or the second one?"*

This illustration captures the probabilistic way the model thinks of language and its own process. When it started to produce its answer, it predicted that it ought to start with "The." Now, neither we nor the model know how it would continue the sentence. It read our question and its partial answer and then made a prediction. Given the context and the vast corpus on which it sits, what should have come next—second or first? It concluded that "second" makes more sense. And once second is produced, the rest of the answer follows.[133]

Generative interpretation in this simple case thus offers courts a better sense of the relevant probabilities if the parties were intending to use English in its most public and common sense. And it does so without reference to singular, perhaps idiosyncratic, illustrations pulled from the golf course. Of course, it's possible that in the context of their deal, extrinsic evidence pointed to a private meaning—or perhaps trade practice could have pushed the court away from the meaning that the model suggests is normal. And, as we'll discuss, knowing that the court would use the model might have motivated both parties to not so quickly assume that their meaning was unambiguously correct.

---

[133] The usual LLM caveats apply, and the probabilities shouldn't be interpreted literally. The model could, for example, continue the sentence with "The first filing would not control."

C.                    *The Ambiguity Problem*

As *Famiglio* illustrated, the question of whether a term is ambiguous, permitting extrinsic evidence or not, can be outcome determinative. That's true for interpretative methods of all stripes. Even the most free-spirited contextualists are not *that* free. They will not waste the parties' time on a lengthy trial when they think that the language in the contract is simply not "reasonably susceptible" to the interpretation proffered by one of the parties. As a result, a key question in contextualist jurisdictions is which interpretation, exactly, the language is reasonably susceptible to.

Take the well-known case of *Trident v. Connecticut*, often listed as a primary argument against California-style textualism.[134] A group of lawyers, assisted by other real estate investors, sought to buy commercial real estate to build their law offices. They borrowed $56 million from Connecticut Insurance, with an agreement to pay it back over 15 years at 12.25% APR. At one point, the agreement stated that the principal could not be prepaid, at least not within the first 12 years of the agreement. However, interest rates fell, and the borrowers sought to prepay the loan with money they would borrow elsewhere.[135] When they were rebuked, they turned to litigation.

The promissory note clearly stated that the borrowers "shall not have the right to prepay the principal amount hereof in whole or in part." But they pointed to a different clause, creating a 10% prepayment penalty for defaulted loans if the lender accelerated. The borrowers' lawyers relied on a famous statement of California's contextualism rule, *Pacific Gas,*[136] to argue that they ought to be permitted to offer extrinsic evidence— negotiations, trade usage—in support of their contractual reading.[137]

In the Ninth Circuit, Judge Kozinski used the case to offer what others have described as a "shrill attack" on the looseness of the California parol evidence rule.[138] He discounted the borrower's prepayment argument, since it was at the lender's option. And he concluded that the contract's "shall not have the right" clause was crystal clear that prepayment was forbidden—standing alone, it was not reasonably susceptible to the borrower's meaning. Nonetheless, Judge Kozinski remanded the case. He wrote:

> Under *Pacific Gas,* it matters not how clearly a contract is written, nor how completely it is integrated, nor how carefully it is negotiated, nor how squarely it addresses the issue before the court: the contract cannot be rendered impervious to attack by parol evidence. If one side is willing to claim that the parties intended one thing but the agreement provides for another, the court must consider extrinsic evidence of possible ambiguity. If that evidence raises the specter of ambiguity where there was none before, the contract language is displaced and the intention of the parties must be divined from self-serving testimony offered by partisan witnesses whose recollection is hazy from passage of time and colored by their conflicting interests . . .

---

[134] 847 F.2d 564.

[135] Historic rates had fallen by around 3 percent, meaning an early pre-payment would have meant a saving of ~$1.1 million over the life of the loan.

[136] Pacific Gas & Electric Co. v. G.W. Thomas Drayage & Rigging Co., 69 Cal.2d 33, 442 P.2d 641 (1968).

[137] 847 F.2d at 568(noting reliance on Pacific Gas).

[138] Peter Linzer, *The Comfort of Certainty: Plain Meaning and the Parole Evidence Rule*, 71 FORDHAM L. REV. 799, 805 (2002).

The opinion, written with flair, is in many contracts casebooks, but it is a puzzle in its own right. California's existing rule provided that extrinsic evidence was to be admitted only if the language in the contract was "reasonably susceptible" to the interpretation proffered by the parties. Thus, if Kozinski really had been confident that the language was clear, he should not have remanded.[139]

Perhaps Kozinski should have the courage of his convictions. After obtaining the original promissory note,[140] we introduced the relevant parts to four models: GPT-4, Turbo GPT 3.5, Claude 2, and Claude 1.3, and then asked for their evaluation. We asked them to read the entire contract and then estimate, as a judge, the likelihood that the parties intended early repayment to be permitted under the agreement. To capture a range of model responses, we repeated the same question many times, while altering the "temperature" at each step (as we described above).

---

[139] Susan J. Matin-Davidson, *Yes, Judge Kozinski, There Is A Parol Evidence Rule in California—The Lessons of a Pyrrhic Victory*, 25 S.W.U. L. REV. 1, 18-20 (1995). As Prof. Matin-Davidson points out, after remand the defendants won a summary judgment motion and their attorneys' fees. There never was a trial. *Id.* at 4, n.22.

[140] We thank Prof. Todd Rakoff for providing it from his collection.

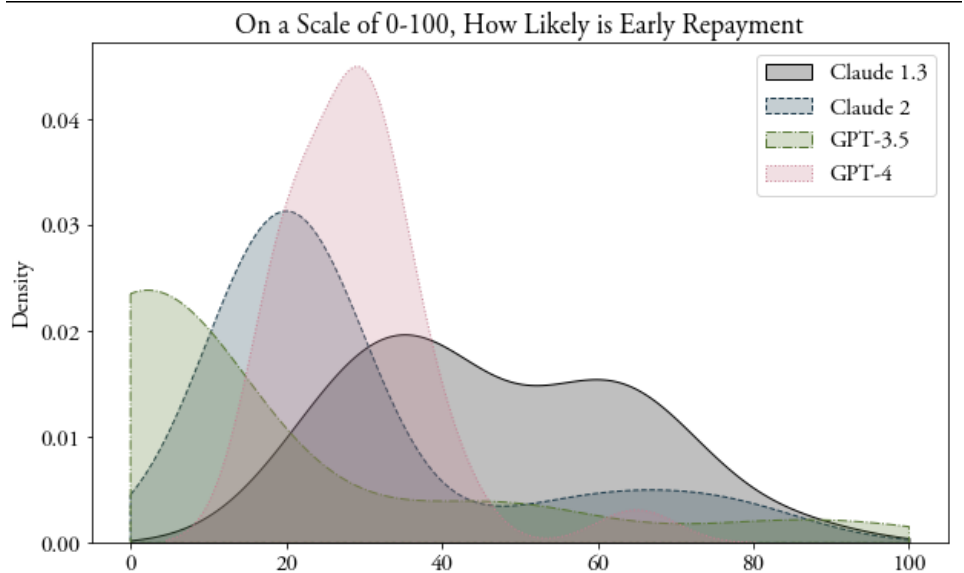

*Figure 4: Turbo GPT 3.5, GPT-4, Claude 1.3, and Claude 2, with 25 varying temperature weights from 0.01-0.2 to 1, fed with Trident facts. On the x-axis, 0 is not very likely and 100 is very likely.*

Figure 4 is suggestive of how generative interpretation can deepen and enrich judicial analysis. Overall, the models all roughly agree that prepayment is not allowed. The least powerful model here, Claude 1.3, was more open to the possibility than GPT 3.5, the model that powers ChatGPT. But the two most powerful models, Claude 2 and GPT-4, both shared a similar evaluation: contra Kozinski, the contract was not very susceptible to the interpretation advanced by the Trident group.

One read of this result is that it suggests that Kozinski's intuitive factual premise was wrong, but that he reached the right conclusion. That is, even taking the borrower's argument seriously, the dominant reading rejects a finding of ambiguity. No further extrinsic evidence ought to have been admitted. This would align with common criticisms of the opinion.[141] On the other hand, the models were *not* uniform in their assessment; the probability distribution suggests that at least *some* probabilistic readings of the contract permit early repayment. To determine the case, we would want to know more about those minoritarian readings: are they reflective of discrete linguistic communities, private meanings, or other legally relevant factors? Generative interpretation does not answer the question of whether language is reasonably susceptible to a meaning, it instead

---

[141] *See* Matin-Davidson, *supra* note 139.

helps us visualize a broad spectrum of meaning and quantify how likely a particular result is.[142]

Now consider another case turning on ambiguity: *Ellington v. EMI*.[143] The issue in this case arose from a 1961 net receipts agreement between the musician Edward Kennedy "Duke" Ellington and his record company, EMI. As was common at the time the parties agreed on a 50/50 royalty split, after deducting fees charged by third parties that intermediate in foreign markets. This net receipt agreement bound EMI and its "other affiliates." In the intervening decades, the music industry underwent significant consolidation, and EMI began to use its own affiliates rather than rely on third parties for foreign operations. It sought to deduct those affiliate fees before paying Ellington's estate.

Feeling blue, Ellington's grandson sued, arguing that two key phrases in the contract were ambiguous: "(1) the phrase "net revenue actually received" in the royalty provision and (2) the term "any other affiliate" in the definition of Second Party."[144] The New York Court of Appeals – the country's preeminent textualist tribunal—rejected the claim. The majority held that the terms were unambiguous: they only reference affiliates that existed at the time of contracting. There is simply no way that they could be read in any other way, given the tense that the parties used and the Court's aversion to forward-looking language.[145]

Again we had access to the original contract. We presented it to GPT-4 for plain language analysis, asking: "does 'other affiliates' naturally include only the existing affiliates at the time of contract, or does it potentially encompass affiliates that might be created over time?"

Before we describe the model's answer, we should highlight two robustness concerns with model interpretation. Models are quite sensitive to the prompt used. This opens them to a problem of "leading prompts," queries that lead the model towards a desired answer. And, as we described earlier, models can be set to be hotter (more random) or colder (more deterministic). This allows the user (judge, researcher, policymaker) many degrees of freedom.

---

[142] When a conclusion that is 20% likely is legally *reasonable* might turn on several factors we do not explore in the text. Imagine a particular linguistic subcommunity whose understanding terms correlate with the parties. (You could think of this as akin to trade usage, but for culture.) In that case, deferring to majoritarian readings would tend to suppress important perspectives. See generally Dan Kahan, David A. Hoffman and Don Braman, *Whose Eyes are You Going to Believe: Scott v. Harris and the Perils of Cognitive Illiberalism*, 122 HARV. L. REV. 837 (2009) (discussing how simulations can uncover discrete minority perspectives on legally-operative facts that the law should attend); David A. Hoffman, *From Promise to Form: How Contracting Online Changes Consumers*, 91 N.Y.U. L. Rev. 1595 (2016 ) (arguing that younger parties have distinct views of contracting than older ones).

[143] Ellington v. EMI Music, Inc., 24 N.Y.3d 239, 241, 21 N.E.3d 1000, 1001 (2014).

[144] *Id.* at 245.

[145] *Id.* at 246-47.

To deal with these issues we tried something new. Rather than a single prompt, we used 20 variations of the same question, each measured at 10 different temperatures.[146] We presented yes/no questions where yes indicates agreement with the judge's interpretation. The Figure below summarizes the results of the experiment among four of the leading models.

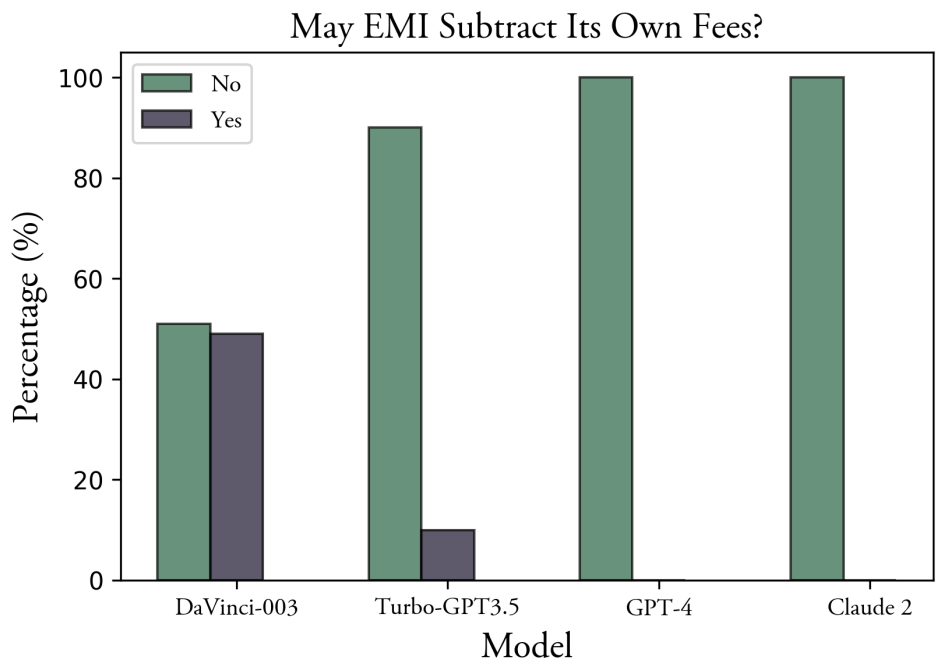

*Figure 5:Ellington v. EMI, analyzing the interpretation of "other affiliates" using 10 equally spaced values between 0.01 and 1, responding to 20 prompt variations generated by GPT-4, after seeding it with the background of the case.*

As the Figure illustrates, the two models don't share the New York court's confidence: the most common interpretation of "other affiliates" includes those that post-date the contract. But we see some interesting diversity among the models. This is likely due to the fact that EMI's argument does have facial plausibility. Of course, even coherence within more powerful models proves nothing. It is illustrative of the value of the method as a convenient check against overconfidence, and a spur to greater reflection. (Though the fact that the dissent thought that the contract was ambiguous might have produced that same introspection.)

---

[146] Specifically, we use 10 equally spaced values between 0.01 and 1. At very high temperatures, the model loses coherence, so this is a conservative range. The 20 prompts were generated by GPT-4, after seeding it with the background of the case and a seed question. AMBIGUOUS CONTRACT INTERPRETATION, https://chat.openai.com/share/e9003c92-5e32-436c-816c-c2add7ac485b (last visited July 29, 2023). For code, *see supra* note 21.

## D. Filling Gaps

One of the most difficult issues in contract interpretation is distinguishing silence from unexpected gaps. Contracts are incomplete: the parties leave many topics to necessary implication. Such omissions are not always deliberate: sometimes parties simply have not contemplated a problem—a global pandemic, a supply chain disruption, another peerless ship sailing ex Bombay[147]—and the Court must engage in filling the gaps, rather than merely interpreting words on the page.[148]

Consider the 1977 New York Court of Appeals case, *Haines v. City of New York*. It resolves a dispute about a 1924 contract between the City of New York and an upstate village, in which the City promised to pay the town to process its own sewage so that the City's water supply could be cleaned. (That is, the City paid the village not to pollute.) As the decades passed, the townships grew and the Federal Government passed environmental regulation. By the early 1970s, facing strong budgetary pressures, New York refused to continue to pay for the township's expansion of the sewage facilities. A local developer sued, arguing that the contract's absence of a duration term or cabin on the scope of the City's obligation meant that the City was in breach.

The Court considered those arguments in a decision that looked only to the written contract. It determined that the parties did not mean for the contract to run forever, in a provision notable for its brevity.

> [W]here the parties have not clearly expressed the duration of a contract, the courts will imply that they intended performance to continue for a reasonable time but also did not mean it to be terminable at will . . . Thus, we hold that it is reasonable to infer from the circumstances of the 1924 agreement that the parties intended the city to maintain the sewage disposal facility until such time as the city no longer needed or desired the water, the purity of which the plant was designed to insure.[149]

---

[147] Raffles v. Wichelhaus, 159 Eng. Rep. 376 (1864).

[148] On this generally—and expressing useful skepticism about the borders—*see* Klass, *supra* note 40; see also Larry Solum, *Legal Theory Lexicon: Interpretation and Construction*, LEGAL THEORY BLOG (May 31, 2020), at https://lsolum.typepad.com/legaltheory/2020/05/legal-theory-lexicon-interpretation-and-construction.html (explaining the difference).

The logic here isn't compelling, but is clear enough: by default, parties do not intend contracts to be terminable at will when they write unlimited obligations, and nothing about the language or circumstances of the contract compels a contrary conclusion.

On the related question of whether the City promised (implicitly) to continue to expand the system's capacity, the Court was less generous.

By the agreement, the city obligated itself to build a specifically described disposal facility and to extend the lines of that facility to meet future increased demand. At the present time, the extension of those lines would result in the overloading of the system. Plaintiff claims that the city is required to build a new plant or expand the existing facility to overcome the problem. We disagree. The city should not be required to extend the lines to plaintiffs' property if to do so would overload the system and result in its inability to properly treat sewage. In providing for the extension of sewer lines, the contract does not obligate the city to provide sewage disposal services for properties in areas of the municipalities not presently served or even to new properties in areas which are presently served where to do so could reasonably be expected to significantly increase the demand on present plant facilities.[150]

Once more the court alludes to the agreement, but its decision is inattentive to the details. It found an implicit condition to obligation: extension is required only so long as the system is not overloaded.[151] But this was a gap-filling exercise, informed by the court's judgment about what the parties *should* have said.[152] Such determinations were part of a waning trend in New York courts who abandoned classical adherence to express terms in favor of a looser, Cardozian approach to missing terms.[153]

---

[149] Haines, 41 N.Y. 2d 769 at 772 (breaks added, cleaned up).

[150] *Id.* at 773.

[151] The City of New York at the time was under severe financial stress and courts rushed to protect it from bankruptcy. Robert M. Jarvis, Phyllis G. Coleman and Gail Levin Richmond, *Contextual Thinking: Why Law Students (and Lawyers) Need to Know History*, 42 WAYNE L. REV. 1603, 1613 (1996).

[152] For an argument suggesting that there is no fact-of-the-matter about parties' intent when filling gaps in contracts, see Robert A. Hillman, *More Contract Lore*, 94 TUL. L. REV. 903, 910 (2020); Robert A. Hillman, *The Supreme Court's Application of "Ordinary Contract Principles" to the Issue of the Duration of Retiree Healthcare Benefits: Perpetuating the Interpretation/Gap-Filling Quagmire*, 32 ABA J. LAB. & EMP. L. 299, 320 (2017).

With the cooperation of the New York court system, we obtained the 1924 contract.[154] This contract and the various exhibits are long, especially considering when they were created: about 8 pages of Word documents. We entered the text into the two models that can support such long inputs—GPT-4's experimental version and Claude 2—and asked them to assess the validity of several legal arguments given the agreements.[155] Figure 6 illustrates what we found.

---

[153] Perhaps this part of the opinion responded to the City's financial exigency. William E. Nelson, *A Man's Word and Making Money: Contract Law in New York, 1920-1960*, 19 MISS. COLL. L. REV. 1, 13 (1998).

[154] E-mail from Marisa Gitto, Reference Services, New York State Library, to Michael Hurley, Research Assistant, University of Pennsylvania Carey Law School (May 22, 2023, 03:01 EST) (on file with authors).

[155] GPT-4: https://poe.com/s/Vp9tkyhGnMmHqFvdKp4n You should take model's self-reported degree of confidence with a grain of salt; it is more meaningful to simply compare its expressed confidence with respect to different questions, hence our experiment design here.

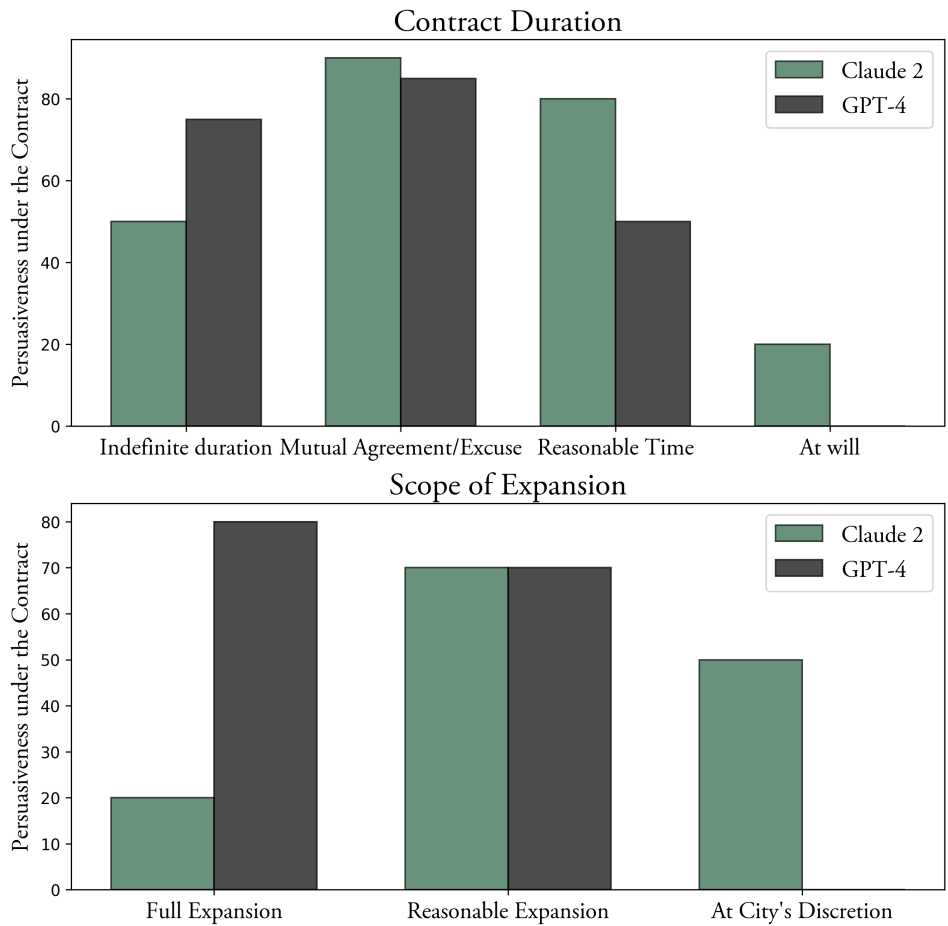

*Figure 7:Haines v. City of New York gap filling analysis using Chat GPT-4 (32k context length) and Claude 2 (100k context length).*

The first set of questions concerned duration. Both models reject the city's claim that the contract was terminable at will. And both (with different degrees of confidence) were open to durational gap fillers of an indefinite time, a reasonable time, by joint agreement or when a legal excuse is present—which is indeed the common law rule for most contracts. GPT-4 (like the court) explained "while a reasonable duration might be inferred under common law principles, this argument does not strongly accord with the contract's language."[156] Overall, the models appear to generally support the court's reading.

---

[156] Likewise Claude-2 explained: "A reasonable duration could be implied, though not explicitly stated."

The second set of questions involved the scope of the City's obligations. GPT-4 disagreed strongly with the court; it thought that the City's obligation was unbounded. Importantly, it anchored its reasoning in a section neglected by the court: Section 6. That part obligates the city to extend sewage plans "[w]henever extensions of any of the sewer lines are necessitated by future growth . . . of the respective communities." For ChatGPT-4, this provision implied the obligation to build additional treatment plants. But Claude 2 was more amenable to the court's interpretation and provided a plausible constraining argument: "The agreement provides for extensions when required by growth, implying a reasonable obligation."

### E. From Text to Context

So far, we have provided examples that showcase how large language models might power a stronger, cheaper, more robust form of textualism. We now consider how such models can account for contextual evidence such as prior conversations, shared expectations, and industry standards. *Stewart v. Newbury* provides a simple illustration.[157] In *Stewart*, a contractor and a business corresponded about the construction of a new foundry. The contractor's offer letter was brief; he offered to do the job and charge either by offering an itemized list or by charging on a cost + 10% basis. This letter was followed by a telephone call where they may have agreed that payment would be made "in the usual manner." Finally, the foundry responded in writing that, following the phone conversation, they accepted the bid. As far as we know, that amounts to the entirely of the contracting case file.[158]

Once the contractor finished the first part of the project, he submitted a bill. The foundry refused to pay. The contractor insisted that it was customary to pay 85% of payments due at the end of every month, but the foundry argued that its payments were only due on (substantial) completion of the project. Seeing no payments made, the contractor stopped work. The parties countersued for breach.

Today, the default rule is that payments in construction contracts are not due until the contract is substantially performed.[159] It is unclear that this rule was in place when the parties agreed in 1919. The foundry argued that no payment was due under the contract, and hence, the contractor's refusal to work was wrongful. So now we have an interpretive question: did the parties agree to override this default?

---

[157] 220 N.Y. 379 (1917).
[158] *Id.* at 380-84.
[159] *See* 22 N.Y. JUR. 2D CONTRACTS § 352, Hillman, *supra* note 152, at 313 ("courts in construction cases find a duty to pay only after substantial performance").

The written agreement is too sparse to help, but the phone conversation offers an in. If we believe that the parties indeed agreed to make payments *in the usual manner*, then it is possible to interpret *usual* as referring to an alleged common practice of monthly installment payments. It is also possible, however, that 'usual' refers to other standard payment conventions—say, the payment on a cost +10% basis.

The court remanded because of faulty jury instructions, so the interpretative question was left undecided. We, however, are not so constricted. We asked today's leading LLM models, GPT-4 and Claude-4, to predict what the parties meant. To do so, we first told the models to assume that the default legal rule would be that payment is conditioned on substantial performance.[160] Then, we asked the models to estimate how the parties would have interpreted their deal absent consideration of either extrinsic evidence of the phone conversation or evidence of industry norms. We then added the evidence of the phone conversation, to see how the model's confidence changes, and finally, we added evidence of the custom in the industry. Table 1 summarizes the results:[161]

| Does the Owner have to pay monthly? (Instead of after substantial performance) | | |
|---|---|---|
| | GPT-4 | Claude-2 |
| Letter contract alone | 10% | 10% |
| +Phone call | 75% | 20% |
| +Industry Norm | 95% | 90% |

*Table 2: Expressed confidence in "the duty to pay is monthly" based on legal and transactional context. Presented to GPT-4 (32k context window) and Claude-2 (100k context window).*

Table 1 demonstrates how each additional piece of evidence alters the analysis. And for purposes of this case, it shows that, for the models at least, extrinsic evidence was materially important to the outcome.

Illustrating the additional value of each piece of evidence to the decisionmaker is useful in part because it permits a real focus on the *quality*, not just the *weight*, of evidence. Did this conversation really transpire in the way alleged by the contractor? Was sufficient evidence proffered to substantiate the industry custom alleged by the contractor? And, later, whether interpreting the contract in this way is socially beneficial?[162] The model can give structure to the evaluation of extrinsic evidence. And within the limits of its prompts, its conclusions are coherent, cheap, and seemingly plausible.

---

[160] This is not obviously the correct legal rule, then or now, but we had to start somewhere, and we took the court at its word.

[161] CLAUDE 2.0 POE CONSERVATION, https://poe.com/s/wLkeCDrPdFpKye3uApSa (last visited July 30, 2023). Again, you should be skeptical of model's expressed confidence; the direction of change with every new piece of evidence, not its quantification, is reliable.

## III.    THE FUTURE OF CONTRACT INTERPRETATION

So convenient are today's LLMs, and so seductive are their outputs, that it would be genuinely surprising if judges were not using them to resolve questions of contract interpretation as we write this article, only a few months after the tools went mainstream. Looking at practical guidance offered to lawyers in the summer of 2023, we see lawyers are encouraged to use LLMs to perform legal research, draft deposition questions and contracts, and predict settlement values.[163] And there are hints that judges are already using ChatGPT to answer other kinds of interpretative questions, just as they would use Google.[164] In one recent survey, one-quarter of judges confessed to using the tool, though many expressed concern about its reliability.[165]

These models are useful because they offer tools—fast, cheap, sometimes incorrect—in service of old interpretative goals. Courts will soon take a phrase like "dozen" and ask ChatGPT to interpret it, rather than turning to the dictionary or Google; or ask the model what's the likely assumption a contract makes when it leaves a gap; or check if it thinks an insurance policy contemplated deft burglars. They'll do so both covertly and overtly, both *sua sponte* and in response to briefing. Almost certainly the first briefs to affirmatively argue for the use of the tool will come from resource-constrained firms. As we illustrated in Part II of this Article, LLMs are already applicable to live problems that courts face every day, and it would be naïve to think they aren't using them.

---

[162] Williston suggests that it is more difficult to order specific performance against a recalcitrant contractor than to order (effective) specific performance against a non-paying principal. 15 WILLISTON ON CONTRACTS § 44:48 (4th ed.).

[163] Catherine Casey, Reveal Brainspace, Ronald J. Hedges, Ronald J. Hedges LLC, Marissa J. Moran, N.Y.C. Coll. of Tech., Stephanie Wilson, Reed Smith LLP, *Generative Artificial Intelligence in Practice: What It Is and How Lawyers Can Use It* (June 28, 2023) (on file with authors).

[164] Luke Taylor, *Colombian Judge Says He Used ChatGPT in Ruling*, THE GUARDIAN (Feb. 2, 2023, 9:53 PM), https://www.theguardian.com/technology/2023/feb/03/colombia-judge-chatgpt-ruling.

[165] Ed Cohen, *Most Judges Haven't Tried ChatGPT, and They Aren't Impressed,* THE NAT'L JUD. COLL. (July 21, 2023), https://www.judges.org/news-and-info/most-judges-havent-tried-chatgpt-and-they-arent-impressed/.

Indeed, we've seen this story play out many times before—as some readers will recall, when courts first realized that Wikipedia could be used as a source of information,[166] they were chastised for its use by higher courts,[167] and then it was eventually folded into the normal set of legal research tools.[168] But at least in the short run, judges won't have the tool draft opinions. And why would they? That courts are irreducibly part of the interpretative enterprise—no matter how sophisticated prediction machines get—follows from the obvious point that there are two stages to every contract interpretation problem: figuring out what the parties meant (at contracting) and deciding the "legal significance that should attach to the semantic content."[169] The method is simply better for many reference purposes than those currently on offer.

The problem then is not *whether* courts will use LLMs as an aid to interpretation, but *how*. Generative interpretation is a tool and as such, it has strengths, limits, and flaws. To be sure, AI's most enthusiastic wielders will be its least careful adopters. Thus, our goal in Section A is to delimit some principles and limitations for its usage by lawyers and judges. With the proper usage of the tool in mind, in Section B we suggest that generative interpretation has implications for the continuing vitality of longstanding debates between textualism and contextualism. Or to put it differently, while the uses that we suggest in Part A could be thought of as Textualism 2.0—better dictionaries and canons—we don't think that's the practical limit of what this method of interpretation can do.

    *A.*          *Interpretation for the 99%?*

As we've said, in the coming months and years, we're sure you will read examples of lawyers and judges using ChatGPT and related tools in perverse, sometimes outright silly ways, and reaching absurd results you think would have been avoided had they just buckled down and done their jobs like careful jurists ought to. Or, worse, they'll have these tools generate pedestrian prose that looks like soulless briefing or opinion-writing, but in fact is built on a throne of lies. There's no question that AI will sometimes be a crutch for lazy or harried lawyers who simply didn't focus on the details: it might not be ideally pitched at the kinds of people who are reading sentences with care 20,000 words into a law review article.

---

[166] Lee F. Peoples, *The Citation of Wikipedia in Judicial Opinions*, 12 YALE J. L. & TECH. 1, 28 (2010) ("Citations to Wikipedia entries in judicial opinions have been steadily increasing since the first citation appeared in 2004.").

[167] Campbell ex rel. Campbell v. Sec'y of Health & Hum. Servs., 69 Fed. Cl. 775, 781 (2006) ("rejecting special master's reliance on Wikipedia, among other online sources, citing several "disturbing" disclaimers on the website and that it could be edited by "virtually anyone"); *see also* Kenneth H. Ryesky, *Downside of Citing to Wikipedia*, N.Y. L.J., Jan. 18, 2007, at 2.

[168] Jodi L. Wilson, *Proceed with Extreme Caution: Citation to Wikipedia in Light of Contributor Demographics and Content Policies*, 16 VAND. J. ENT. & TECH. L, 857, 907 (2014) ("The advent of Wikipedia and other technological advances has changed legal research. It is unrealistic to believe that the legal community can ignore that reality. . . .").

[169] Schwartz & Scott, *supra* note 34, at 568 n.50; *see generally* Edwin W. Patterson, *The Interpretation and Construction of Contracts*, 64 COLUM. L. REV, 833, 833–35 (1964); Klass, *supra* note 40.

And yet it's precisely because LLMs are cheap and workmanlike that they will be of real use in contract interpretation. The biggest single problem with all currently available approaches to contract interpretation isn't that they are incapable of getting correct results some of the time. It's that they are inaccessible to ordinary parties.[170] The result is that non-wealthy individuals who suffer breach have to lump it,[171] tilt against corporations in internal dispute resolution systems,[172] or face financially ruinous fees and prevail in pyrrhic victories.[173] Simply put: there is an access-to-justice problem at the center of contract law as pernicious as the better recognized ones in criminal and constitutional adjudication. The costs and uncertainties of interpretating deals, which form the core of contract litigation, materially contribute to this problem.[174]

Costly interpretation burdens judges too. Chambers are not endowed with reference experts on call for every query. Courts have fewer resources and competencies than the layperson would imagine. This stylized fact alone can explain why dictionaries are popular, and why corpus linguistics is at best experimental; why law office history exists but not law office econometrics; and perhaps even why federal precedent on state issues is more cited than the relevant state law, given that the former is thoroughly indexed in common commercial databases and the latter is not.[175] To substitute for dictionaries and familiar Latin canons, new interpretative tools must be free (or nearly so) and widely available. LLMs satisfy those conditions. Already today, interactions through a chat interface do not require more skill than using a search engine. The deft burglar example offers a proof of concept, and the remaining examples (though not immediately available in your chatbot window) are likely months, not years, away.

---

[170] SEE LEGAL SEVS. CORP., THE JUSTICE GAP: MEASURING THE UNMET CIVIL LEGAL NEEDS OF LOW-INCOME AMERICANS 6 (2017) ("86% of the civil legal problems reported by low-income Americans in the past year received inadequate or no legal help."); E.H. Geiger, *The Price of Progress: Estimating the Funding Needed to Close the Justice Gap*, 28 CARDOZO J. EQUAL RTS. & SOC. JUST. 33, 34-39 (2021) (documenting an array of causes behind the "justice gap").

[171] Geiger, *supra*, at 38 ("[T]he average household faces 9.3 legal issues per year. 65% of those problems are never resolved; potentially because the claimants cannot afford counsel and do not have the legal literacy to pursue their claims pro se.").

[172] *See* Rory Van Loo, *The Corporation as Courthouse*, 33 YALE J. REG. 547 (2016).

[173] Matthew R. Hamielec, *Class Dismissed: Compelling a Look at Jurisprudence Surrounding Class Arbitration and Proposing Solutions to Asymmetric Bargaining Power Between Parties*, 92 CHI.-KENT L. REV. 1227, 1231 (2017) (arguing that class action waivers and arbitration provisions can result in "negative value suits" where low-resource claimants are pitted against wealthier opponents); Gideon Parchomovsky & Alex Stein, *The Relational Contingency of Rights*, 98 VA. L. REV. 1313, 1340 (2012) (noting that class actions can transform individual negative value suits into a single positive value action).

[174] Ben-Shahar & Strahilevitz, *supra* note 18, at 1757-58 (discussing interpretation costs); CATHERINE MITCHELL, INTERPRETATION OF CONTRACTS: CURRENT CONTROVERSIES IN LAW 110 (2007) (noting expenses associated with contextual approaches to interpretation).

Generative interpretation is a tool which responds to this access-to-justice concern. If courts commit to the method, the costs of achieving accuracy in contract interpretation disputes will fall.[176] That's so because the less precise, even if relatively cheap, forms of textualist evidence—dictionaries and canons—will be replaced by better ones. As dispute costs fall, and outcomes become more predictable, the returns to opportunistic breach, which generally benefits sophisticated players, will fall.[177] Over time, one possibility is that there will be *fewer cases* to adjudicate, because parties will likely have a much better sense of what they'll get at verdict, and settle accordingly.[178] And better calibrated results *ex post* means that parties can spend less time (and money) contracting *ex*

---

*ante.*[179] A promise of generative interpretation—which it may yet fulfill—is that it will open a form of textualism up to the 99%.[180]

The pages of law reviews are littered with proposed technological solutions to supposed problems of excessive legal costs, which turn out to be either more intractable than the authors thought or ignore virtues that the authors discounted. We should proceed with care, especially when recommending the widespread adoption of a chatbot that sits on matrices whose outputs even its creators do not well-understand. The question is not (in our view) whether generative interpretation offers predictions that are superior in all cases to artisanal, careful, linguistic analysis. It's whether the method is *good enough,* right now or soon, for resource-deprived courts to adopt in ordinary cases. In evaluating that question of basic competency, it's meaningful that even today's unspecialized models can replicate the results of well-considered cases (as Section II explored) and prompt courts to consider their own priors.

But Section II offered a curated tour of generative interpretation's greatest hits. It didn't show you where things can go wrong. To make this tool perform as well as it can, users should be cognizant of these issues and use it according to evolving best practices. To begin, let's start with hallucinatory outputs.[181] In a now-famous case from May 2023, lawyers in a New York Federal court turned to ChatGPT for help researching a motion. The tool obliged with helpful cites, but unfortunately had completely made up the opinions in question.[182] A sanctions order and plenty of bad press followed.[183] In response to the case, other judges have required lawyers to certify that they had not used any form of Artificial Intelligence in their filings.[184]

---

[179] *See* Spencer Williams, *Predictive Contracting*, 2019 COLUM. BUS. L. REV. 621 (arguing that parties could use information about contract outcomes, harnessed through machine learning of large datasets, to change out they contract ex ante). But for an insightful discussion of how selection operates to make difficult machine predictions about litigation outcomes, *see* David Freeman Engstrom and Jonah Gelbach, *Legal Tech, Civil Procedure, and the Future of Adversariliasm*, 169 U. PA. L. REV. 1001, 1065-1067 (2021) (discussing obstacles to prediction).

[180] Schwartz & Scott, *supra* note 34, at 941 (noting "the more time the court spends on a particular interpretive issue, the less time it can spend on other issues or other cases").

[181] Sharon D. Nelson, John W. Simek & Michael C. Maschke, *Beware of Ethical Perils When Using Generative Ai!,* 46-JUN, WYO. LAW., at 28, 30 ("In fact, it can come up with very plausible language that is flatly wrong. It doesn't 'mean to' but it makes things up--and that is what AI researchers call a 'hallucination'. . . .").

[182] Benjamin Weiser, *Here's What Happens When Your Lawyer Uses ChatGPT*, THE NEW YORK TIMES (May 27, 2023), https://www.nytimes.com/2023/05/27/nyregion/avianca-airline-lawsuit-chatgpt.html.

[183] *See* Mata v. Avianca, Inc., __ F. Supp. 3d. __, 2023 WL 4114965 (June 22, 2023).

False outputs arise from the predictive nature of generative models.[185] Hallucinations are generated texts asserting facts that are not quite true.[186] Large language models, remember, are statistical tools optimized to make predictions. But LLMs are not like a helpful librarian that simply pulls out the most relevant book on a topic. Facts are stored in the LLM similar to the way other reasoning and statistical facts are stored, as floating points in a labyrinthian array of vectors. When asked to provide a source on a legal matter, the model employs the same method to elicit both facts and inferences. The output doesn't distinguish facts from inferred facts, and sometimes will predict the world incorrectly.

Recent work has made significant advances in understanding and mitigating hallucination errors, and more powerful models are less susceptible.[187] One solution that is already used in some contexts is connecting the model to a database of facts, so that it can act more like the helpful librarian.[188] So while it is appropriate to pay attention to the hallucination problem, it's not obvious that it is a fundamental, persistent, and broad concern. That said, as a best practice, judges would do well to cross-verify the answers that they get from one platform against another, just as in the early days of legal research it would pay to check both Lexis and Westlaw to make sure that your research was complete.[189]

---

[184] Devin Coledwey, *No ChatGPT in my court: Judge orders all AI-generated content must be declared and checked*, TECHCRUNCH (May 30, 2023, 7:32 PM), https://techcrunch.com/2023/05/30/no-chatgpt-in-my-court-judge-orders-all-ai-generated-content-must-be-declared-and-checked/ (explaining the order, which states that "no portion of the filing was drafted by generative artificial intelligence (such as ChatGPT, Harvey.AI, or Google Bard) or that any language drafted by generative artificial intelligence was checked for accuracy, using print reporters or traditional legal databases, by a human being.").

[185] Benj Edwards, *Why ChatGPT and Bing Chat are so good at making things up*, ARS TECHNICA (Apr. 6, 2023, 11:58 AM), https://arstechnica.com/information-technology/2023/04/why-ai-chatbots-are-the-ultimate-bs-machines-and-how-people-hope-to-fix-them/ ("the model is fed a large body of text . . . and repeatedly tries to predict the next word in every sequence of words. If the model's prediction is close to the actual next word, the neural network updates its parameters to reinforce the patterns that led to that prediction."); waka55 (u/wakka55), REDDIT (Apr. 16, 2023, 2:48 PM), https://www.reddit.com/r/OpenAI/comments/12okltx/openais_whisper_api_sometimes_returns_what_looks/ (showing that this problem is not limited to textual generation).

[186] Beren Millidge, *LLM's confabulate not hallucinate*, BEREN'S BLOG (Mar. 19, 2023), https://www.beren.io/2023-03-19-LLMs-confabulate-not-hallucinate/ (describing problem).

[187] *See e.g.*, Matt L. Sampson & Peter Melchior, *Spotting Hallucinations in Inverse Problems with Data Driven Priors*, ARXIV: 2306.13272 (JUNE 23, 2023), https://arxiv.org/pdf/2306.13272.pdf (arguing that hallucinations can be qualitatively differentiated from fact-based inferences by focusing on activation regions); *see also* Philip Feldman, James R. Foulds, & Shimei Pan, *Trapping LLM Hallucinations Using Tagged Context Prompts*, ARXIV: 2306.06085 (June 9, 2023), https://arxiv.org/abs/2306.06085; *see also* Ayush Agrawal, Lester Mackey, & Adam Tauman Kalai, *Do Language Models Know When They're Hallucinating References?*, ARXIV: 2305.18248 (May 29, 2023), https://arxiv.org/abs/2305.18248; *see also* Gabriel Poesia, Kanishk Gandhi, Eric Zelikman, & Noah D. Goodman, *Certified Reasoning with Language Models*, ARXIV: 2306.04031 (June 6, 2023), https://arxiv.org/pdf/2306.04031.pdf.

Second, models are subject to manipulation. Large language models are malleable; "leading prompts" can lead them to different conclusions. This is roughly analogous to leading questions for witnesses or jury instructions that frame disputes for or against a particular outcome. As anyone who has experience with an LLM chat bot will attest, it is relatively easy to drive conversations toward desired outcomes. In litigation practice, we should expect that the parties themselves will submit competing prompts, just as they vie to control the framing of the legal questions in litigation today. In response, factfinders can (as we illustrated above) ask the model to itself produce competing prompts, and then, rather than relying on a single query, the factfinder can look at the general trend of responses and share those varying outcomes in their decisions.

A third consideration focuses on the models' strength: they are naturally inclined to make predictions that maximize probability—in other words, they are biased towards majoritarian interpretations. Models offer an approximation of general understanding that may simply not be available in any other way, and thus advance long-held goals of contract theory.[190] But majoritarian interpretations are just that: they embed and advance the values of the majority. This is doubly problematic. First, courts really ought to be attentive to local, more private, meanings: public meaning is second best, prioritized because it is efficient and not because it is correct.[191] But more generally, because the linguistic conventions of underrepresented communities are submerged by majoritarian public meanings, they will find it more difficult to have their voices surfaced (and thus subsidized) in contract adjudication. Majoritarian interpretative approaches risk silencing entire communities.[192]

---

[188] *See generally* James Briggs & Francisco Ingham, *Fixing Hallucination with Knowledge Bases*, PINE-CONE, https://archive.pinecone.io/learn/langchain-retrieval-augmentation/.

[189] *See generally* Robert J. Munro, J. A. Bolanos & Jon May, *LEXIS vs. WESTLAW: An Analysis of Automated Education*, 71 LAW LIBR. J. 471 (1978).

[190] Schwartz & Scott, *supra* note 34, at 583-584.

[191] For the foundational work distinguishing local from popular interpretative modes, *see* 2 SAMUEL WILLISTON, THE LAW OF CONTRACTS, § 604, 1162 (1920). Even textualists understand that strict adherence to the public meaning of words, bereft of any commercial understanding of what the parties could have been, will sometimes lead courts astray. *See generally* Stephen J. Choi, Mitu Gulati & Robert E. Scott, *The Black Hole Problem in Commercial Boilerplate*, 67 DUKE L.J. 1, 2 (2017) (describing *pari passu* clauses as "a standard provision in sovereign debt contracts that almost no one seems to understand").

Surely, this is not a problem unique to generative interpretation: dictionaries, canons, and corpora are equally, if not more, vulnerable to the charge.[193] And unlike dictionary-and-canon-textualism, it is at least theoretically possible to query models about the linguistic conventions of distinct communities, enabling courts to come closer to understanding what the parties before them intended their language to mean.[194] This is an active area of research and regulatory scrutiny and should check factfinders.[195]

Fourth, models may become subject to parties' adversarial attacks.[196] By way of illustration, modern AI systems can reliably differentiate between pictures of panda bears and horses, and stop signs and yield signs. But if a sophisticated party can imperceptibly change the color of a pixel here and there, that will be enough to make the model see a horse or a yield sign.[197] The same manipulations can be used to "attack" LLM models.[198] Slight changes in the wording of a contract—e.g., subtle changes in the presentation of the words—might hack the model logic system and alter its interpretation.[199] There is no known general solution to such issues. But if judges and parties become aware of the possibility of such subtle manipulations, they might develop defenses, like using sanitized versions of the contract in their analyses.

---

[192] *See, e.g.,* Majorie Florestal, *Is a Burrito a Sandwich,* 14 MICH. J. RACE & L. 1, 36-39 (2008) (discussing role of race and class in an interpretation dispute); Alexandra Buckingham, Note, *Considering Cultural Communities in Contract Interpretation,* 9 DREXEL. L. REV. 129 (2016) (arguing for the use of cultural meaning in interpretation); *see also supra* note 142.

[193] Steven J. Burton, *A Lesson on Some Limits of Economic Analysis: Schwartz and Scott on Contract Interpretation,* 88 IND. L.J. 339, 350 (2013) (arguing that majoritarian readings can privilege certain views).

[194] For an illustration of this use case, *see* Arbel & Becher, *supra* note 15, at 99-104.

[195] *Proposal for a Regulation of the European Parliament and of the Council Laying Down Harmonized Rules on Artificial Intelligence (Artificial Intelligence Act) and Amending Certain Union Legislation Acts,* at 4, COM (2021) 206 final (Apr. 21, 2021) (stating that a goal of the proposal is to "minimise the risk of algorithmic discrimination, in particular in relation to the design and the quality of data sets used for the development of AI systems. . . .").

[196] For an expanded discussion, *see* Arbel & Becher, *supra* note 15.

[197] Agnieszka M. Zbrzezny & Andrzej E. Grzybowski, *Deceptive Tricks in Artificial Intelligence: Adversarial Attacks in Ophthalmology,* 12(9) J. CLIN. MED. 3266 (2023) ("Suppose we consider even minor perturbations to the image, such as the change in colour of just one pixel. Then, such models are uncertain for small perturbations.").

[198] For a formal exploration, *see* Jindong Wang, Xixu Hu, Wenxin Hou, Hao Chen, Runkai Zheng, Yidong Wang, Linyi Yang, Wei Ye, Haojun Huang, Xiubo Geng, Binxing Jiao, Yue Zhang, Xing Xie, *On the Robustness of ChatGPT: An Adversarial and Out-of-distribution Perspective,* ARXIV: 2302.12095 (Mar. 29, 2023), https://arxiv.org/pdf/2302.12095.pdf.

Fifth, models are sensitive to time. As your neighborhood originalist will tell you, the meaning of words is embedded in the time they were used. If we want to interpret the meaning of a contract signed in 1924, we should account for the linguistic conventions of the time. Models are trained on data indiscriminately: it is unlikely that they will be able to interpret a term as it was read in a specific period in time. The problem is compounded since the training data may include information that was not available for the contracting parties at the time of contracting. This may well include the decision of a trial court when the appellate court seeks to interpret the contract. We can think about this as pollution of the database: for example, perhaps Hurricane Katrina associated levee with flood more closely than it was at the time the relevant insurance contracts were signed.[200]

This problem is longstanding. Judges' innate sense of language is also grounded in the linguistic conventions in which they are personally embedded. Dictionaries and corpus linguistics have an advantage here, because one could seek a dictionary or a corpus from the relevant time period. But even this advantage is limited, because dictionaries are updated in intervals of decades,[201] and corpora cover considerably fewer texts when they are sliced to relevant time periods.[202] Thus, courts will have to consider whether the use of language has shifted over time, and perhaps restrain the use of generative interpretation in cases where its training data suffers from linguistic drift.

---

[199] From the model's perspective, "please" and "please" are not the same word. For an accessible exploration, *see* Computerphile, *Glitch Tokens - Computerphile*, YOUTUBE (Mar. 7, 2023), https://www.youtube.com/watch?v=WO2X3oZEJOA. Various other examples are esoteric: certain models act unexpectedly when presented with specific nonsensical words like "SolidGoldMakigarp." *See* FORBIDDEN TOKENS PROMPTING RESULTS, https://docs.google.com/spreadsheets/d/1PAZNCks11qoUpiojTJpj0od-CYQL2_HGQgam8HSwAopQ/edit#gid=0 (last visited July 20, 2023). But in high stakes settings, such vulnerabilities can be exploited.

[200] A more far-fetched problem is parties trying to inject meaning into the record, just as they would in a normal interpretation dispute by way of after-action lawyer letters and the like. But because parties expect performance, not breach, and the relevant corpora for LLMs is so vast, jurists should worry less about this problem than the internal-to-the-text adversarial attacks we describe above.

[201] *See* HISTORY OF THE OED, https://www.oed.com/information/about-the-oed/history-of-the-oed/?tl=true (last visited July 20, 2023); *See* MERRIAM-WEBSTER ABOUT US ONGOING COMMITMENT, https://www.merriam-webster.com/about-us/ongoing-commitment (last visited July 20, 2023).

[202] Mouritsen, *supra* note 19, at 1378 ("One of the challenges for examining usage in context in a corpus is that the greater the specificity of the search, the fewer examples appear in the corpus.").

Sixth, generative interpretation will need a language of its own. Although scholars often hype objective, scientific methods of proof and judgment, this way of explaining and justifying the exercise of power is uncompelling, and perhaps repulsive, to the population at large.[203] (Which is one reason we've tried to tamp down the statistics and claims to singular answers in this paper.) Juries, after all, aren't presented with simple probabilistic proofs – and judges don't typically justify their decisions by saying they have a 51% chance of being right.[204] Thus, a real problem for the method—which it shares with corpus linguistics and the survey methodologies discussed above—is how to explain itself to lay audiences in ways that reinforce, rather than diminish, judicial legitimacy.[205] It's sociologically normal to say that the word *chicken* takes meaning from the dictionary and trade usage.[206] This sociological framework does not yet exist for black box language models.[207] Courts will have to find ways to wrap the results from automated interpretation in packages that help laypeople to see law as engaging in a values-driven, communal, constrained exercise, and not merely the highest probability next-token predictions.[208]

---

[203] David A. Hoffman & Michael P. O'Shea, *Can Law and Economics Be Both Practical and Principled?*, 53 ALA. L. REV. 335, 339 (2002) ("Most intriguingly, the studies suggest that in certain cases people prefer that legal decisions not be made on an economic basis.").

[204] As Nesson famously argued, the fact-finding system (and juries) exists to achieve legitimacy, not just accuracy. Charles Nesson, *The Evidence or the Event?: On Judicial Proof and the Acceptability of Verdicts*, 98 HARV. L. REV. 1357, 1358 (1985).

[205] *Cf.* Benjamin Minhao Chen, Alexander Stremitzer and Kevin Tobia, *Having Your Day in Robot Court*, 36 HARV. J. L. & TECH. 1 (2022) (presenting experimental evidence that subjects are not biased against algorithmic decisionmakers).

[206] *Cf.* Frigaliment Importing Co. v. B.N.S. Int'l Sales Corp., 190 F. Supp. 116 (S.D.N.Y. 1960) (adopting the broader meaning of the word after contextual inquiry).

[207] Hasala Ariyaratne, *The Impact of Chatgpt on Cybercrime and Why Existing Criminal Laws Are Adequate*, 60 AM. CRIM. L. REV. ONLINE 1, 7 (2023) ("Since ChatGPT uses complex deep learning algorithms, it is often a black box with no clear reason why it provided a certain output."); David S. Rubenstein, *Acquiring Ethical AI*, 73 FLA. L. REV. 747, 766 (2021) ("deep learning neural networks drive some of the most powerful, sophisticated, and functional AI systems, but their complexity renders them inscrutable to humans."); Nelson, Simek & Maschke, *supra* note 181, at 30 ("AI is largely a 'black box'--you cannot see inside the box to see how it works.").

[208] Related to this rhetorical concern is one about attribution and basic fairness that citizens may have about use of LLMs. *See, e.g.*, Sheera Frenkel and Stuart A. Thompson, *'Not for Machines to Harvest': Data Revolts Break Out Against A.I.*, THE NEW YORK TIMES (July 15, 2023), https://www.nytimes.com/2023/07/15/technology/artificial-intelligence-models-chat-data.html; Mark A. Lemley & Bryan Casey, *Fair Learning*, 99 TEX. L. REV. 743, 748 (2021) ("In this Article,

The solution likely lies in a specific type of transparency. Just as much as judges are sociologically committed to certain types of dictionaries, so will it be the case that certain models will emerge as robust and trustworthy. The current practice of interpretation is largely indefensible on this score; because we have no window into the court's processes, we cannot see the dictionaries it did not select or he words it chose not to focus on. But we can know what model a court picks, and from that selection, what probabilities it assessed. We cannot know exactly how the model produced those outcomes, as this knowledge lies in its vast inscrutable matrices. But so long as a judge not only discloses the version of the model that she employed, but also the particular prompts that she used, generative interpretation is more replicable than any other method on offer.[209] (We have tried to show how that would work in the notes of this article.) Indeed, courts might go further: they can *capsule* the results of their inquiries and incorporate them as permanent links to their opinions.

In summary, generative interpretation promises an accessible, relatively predictable, tool that will help lawyers and judges interpret contracts. If it's to achieve that promise, courts will need to be careful to use this tool while being mindful of its uses and limitations. To guide what would inevitably be a process of exploration, we offered a series of best practices based on the technical foundations and legal constraints that define the limits of this tool. As a default, judges should disclose the models and prompts they use and try to validate their analyses on different models and with multiple inputs. Ideally, they'd capsule their findings online. They'll want to be careful about parties' manipulative behavior, and to consider how (and whether) to excavate private, non-majority meanings. By doing so—and by saying what they are doing clearly and with appropriate recognition of LLMs' foibles—courts can fairly experiment with this new technology and achieve a better grasp on the contract's meaning, without abusing the tool or subjecting themselves to reversals.

B.        *Beyond the Textualist/Contextualist Divide*

As we described in part I, the modern debate about interpretation takes as a given that prediction is the goal. But in dividing about how to best accomplish prediction, scholars and courts disagree about an empirical meta-question: how would most parties

---

we argue that ML systems should generally be able to use databases for training, whether or not the contents of that database are copyrighted."); *see also* Peter Henderson, Xuechen Li, Dan Jurafsky, Tatsunori Hashimoto, Mark A. Lemley & Percy Liang, *Foundation Models and Fair Use*, ARXIV: 2303.15715 (Mar. 28, 2023), https://arxiv.org/pdf/2303.15715.pdf.

[209] The model disclosure should include the model's hyperparameters, much like judges share the version of the dictionary they consulted.

prefer that courts interpret their deals?[210] Many have argued that sophisticated parties prefer textualism.[211] Others assert that contextualism is preferred, especially within longer-term relational contexts.[212] Some argue such preferences are, well, contextual.[213] Litigated cases appear to be all over the map.[214] The views of poorer parties are more rarely studied. True, contextualism promises to protect parties from bait-and-switch maneuvers and opportunistic drafting. But who can afford it?

Generative interpretation challenges the utility of this old binary. Starting with textualism, its proponents have said that it builds a common commercial vocabulary and motivates clear contract drafting.[215] But if applied correctly, generative interpretation (as a form of textualism) can predict parties' intent well even without invocation of specialized language or expensive drafting. And if courts follow our proposed best practices, this method is also predictable *ex ante*. When parties can anticipate in advance the choice of model—and we argue that they should be able to contract for it explicitly—then they can clarify disputes well ahead of litigation. Even if the judge consults a broader evidentiary base than the contract itself, models can incorporate it and produce consistent outputs.

By contrast, contextualism promises accuracy by integrating all relevant evidence. Its champions think it protects the weak from the powerful and reflects the real premises of relational contracting relationships.[216] But as a judicial practice, it encourages gamesmanship,[217] exposes decisionmakers to bias-inducing testimonies, increases uncertainty,[218] and more than anything, is simply very expensive. Generative interpretation can also serve as a form of contextualism. It is cheaper to incorporate context into the process when the model can feed on dozens of pages of evidence. Models are not biased by parcels of evidence like human decisionmakers. And armed with LLMs, judges can assess, at the summary judgment stage, the incremental probative value of proposed elements of evidence. As we have demonstrated in *Stewart*, the judge can weigh in advance whether litigation over, say, the records of a phone conversation would be materially important to the outcome. This kind of prioritization is generally the approach of Uniform Commercial Code: the models can turn an interpretation ladder into something more objective.[219]

---

[210] Bayern, *supra* note 93, at 1101.

[211] *See e.g.*, Schwartz & Scott, *supra* note 34, at 941 ("parties prefer textualist interpretive defaults.").

[212] *See* Lisa Bernstein, *Merchant Law in a Merchant Court: Rethinking the Code's Search for Immanent Business Norms*, 144 U. PA. L. REV. 1765, 1769-70 (1996) (business arbitrators avoid business norms); Benoliel, *supra* note 51 (sophisticated parties prefer textualism). For a survey of the scholarly literature, *see* Silverstein, *supra* note 94, at 278-81; *see also* U.C.C. § 2-202(a) (AM. L. INST. & UNIF. L. COMM'N 1951) (usage of trade).

[213] *See* Adam B. Badawi, *Interpretive Preferences and the Limits of the New Formalism*, 6 BERKELEY BUS. L.J. 1, 1 (2009).

[214] Silverstein, *supra* note 52, at 259 (noting courts mixed approaches in litigated cases).

[215] Gilson, Sabel & Scott, *supra* note 49, at 40-41.

[216] *See supra* at text accompanying notes 84 to 93 (discussing contextualism).

[217] Gilson, Sabel & Scott, *supra* note 49, at 41 ("Under a contextualist theory, a party for whom a deal has turned out badly has an incentive to claim that the parties meant their contract to have a different meaning than the obvious or standard one. Such a party can often find in the parties' negotiations, in their past practices, and in trade customs, enough evidence to ground a full, costly trial, and thus to force a settlement on terms more favorable than those that the contract, as facially interpreted, would direct.").

All of this suggests a disruption of the traditional impasse. Generative interpretation allows both predictability and constraint, while also offering better linguistic accuracy.[220] And it corrals litigation costs. Or to put it differently, the choice between four corners or no corners at all is a product of its time and of a specific adjudicatory technology. As this technology improves, judges can relax old safeguards towards a more inclusive approach.

To be sure, generative interpretation would be a simple flip in the default: parties could indicate that their meaning was not to be determined by large language models, just as they can now commit to avoiding certain dictionaries or choosing others.[221] Just as using dictionary to interpret a secret cipher is a foolish way to interpret a deal,[222] following parties' expressed interpretative preferences is wise. Generally speaking, giving parties the ability to control how contracts are interpreted respects their autonomy and carries efficiency benefits.[223] So too here: generative interpretation expands the kinds of evidence that *most* parties would like courts to consider, but it won't be for everyone.

---

[218] Schwartz & Scott, *supra* note 34, at 587; Schwartz & Scott, *Redux, supra* note 34, at 944-47 (arguing that certain parties prefer textualist defaults in part because of the risk of error).

[219] U.C.C. § 1-303(e) (AM. L. INST. & UNIF. L. COMM'N 1977) (order of hierarchy). The hierarchy doesn't always control. See, e.g., Air Prod. & Chemicals, Inc. v. Roberts Oxygen Co., No. CIV.A. 10C12243 FSS, 2011 WL 7063681, at *3 (Del. Super. Ct. Nov. 30, 2011) ("Much like Delaware law, Pennsylvania law prefers the contract's express terms. But, Air Products's course of dealing and course of performance allegations might illuminate the contract and bring its terms into sharper relief.").

[220] Schwartz & Scott, *Redux, supra* note 34, at 946 (suggesting that controlling litigation costs is one reason that sophisticated parties prefer to avoid extrinsic evidence).

[221] 5 MARGARET N. KNIFFIN, CORBIN ON CONTRACTS: INTERPRETATION OF CONTRACTS § 24.9 (Joseph M. Perillo ed., rev. ed. 1998) (courts should enforce private meanings, "however we may marvel at the caprice"); *see, e.g.,* Smith v. Wilson (1832) 3 B. & Ad. 728, 728 (holding that "parol evidence was admissible to sh[ow] that . . . the word thousand, as applied to [the contract], denoted twelve hundred").

[222] KNIFFIN, *supra* note 221, § 24.13 (courts should and do enforce the parties' vernacular).

Even if all generative interpretation does is flip the default on extrinsic evidence surrounding contracting, it still has important distributive effects. Textualism's many virtues can be recast as its elitist faults. Poorer parties, or uncounseled ones, often misunderstand the relationship between contractual disclaimers of reliance and oral sales talk.[224] Though the Restatement of Consumer Contracts suggests that courts should be more open to the idea that contracts that disclaim obligation in the face of contrary promises should not be enforced,[225] it does little to help with interpretative disputes which are less obviously unjust. And yet there are many examples of parties' proffered meaning being excluded as violative of the parol evidence rule,[226] or simply not considered because the meaning is purportedly plain.[227] As the *Ellington* example above demonstrates, even on their own terms such decisions may be questioned. But if the parties have not otherwise indicated, generative interpretation will provide more evidence to courts that extrinsic meaning ought to matter in discerning what the parties contemporaneously would have said they meant.

An exemplary case that generative interpretation could benefit is *Smith v. Citicorp*.[228] The Smiths needed to borrow money to repay an old loan and pay for some home improvements. They turned to Citicorp, which purported to create a revolving loan agreement, secured by their home. The key to the dispute was that the interest rate on this loan was 13.99% APR, a rate only permissible for revolving loans, not closed ones. The Smiths argued that closed is exactly what the loan agreement was. Miraculously, the Smiths had signed affidavits from *two* Citicorp employees, attesting that Citicorp never intended to make advances on this loan (which would have defined an open-ended, revolving, loan). But the Supreme Court of Alabama ignored that highly-probative and rare, evidence because it laid outside the four corners of the contract.

We think this result gives too much weight to generalized worries about courts' competency to evaluate extrinsic evidence. It would be a trivial task to incorporate the affidavits into the generative analysis, and, as we've shown, they can be weighted according to the judge's priors. This would not resolve questions of credibility and relevance, but the flexibility of incorporating it at the margins might radically improve the accuracy of the court's analysis.

---

[223] Schwartz & Scott, *supra* note 34, at 569.

[224] Lawrence M. Solan, *The Written Contract as Safe Harbor for Dishonest Conduct*, 77 CHI.-KENT L. REV. 87, 92 (2001) (identifying ways in which integrated agreements promote injustice).

[225] Restatement of Consumer Contracts § 6 (2023).

[226] Gold Kist, Inc. v. Carr, 886 S.W.2d 425, 430 (Tex. App. 1994), *writ denied* (Mar. 23, 1995).

[227] Greenfield v. Phillies Recs., Inc., 98 N.Y.2d 562, 570 (2002).

[228] Smith v. Citicorp Pers.-to-Pers. Fin. Centers, Inc., 477 So. 2d 308, 311 (Ala. 1985).

Because generative interpretation blurs the line between textualism and methods of interpretation that are more capacious in their evidentiary sources, and because it enables a new set of evaluative metrics and socio-legal advantages, we think that it ultimately won't be (just) Textualism 2.0. Rather, it will become a distinctive method of evaluating contractual meaning, marked by its own jargon, normative commitments, and practitioner community. That new methodology will take time to develop. As we said, in the early days, judges will dip in and out of the application, using it as one would a dictionary, or a refresher CLE on the canons of contract interpretation. Only when lawyers start to argue that the tool can provide better answers to interpretative questions will courts ask if that is true, and whether answers from ChatGPT should supplant those from Merriam's or Black's dictionaries.

## CONCLUSION

In this Article we introduced generative interpretation, a method of interpreting legal texts using large language models. Our work follows a rapidly evolving practice: lawyers and judges are already experimenting with these models in law offices and chambers across the country, some covertly, others less so. We offered a deep dive into the way the technology works (and fails), and explored techniques of using it to better perform interpretive tasks. We demonstrated that the technique can be applied to famous contracts cases, often arriving at the same answers at lower cost and with greater certainty, and while sometimes exposing ambiguities, dislodging sticky priors about meaning, and parceling out the marginal effect of new evidence on interpretation.

In our view, generative interpretation is a tool with important implications for legal practice and contract theory. Because language models are attentive to context, and because they can voraciously digest long texts, they offer a much more robust form of textualism. The model's complex encoding of language far outstrips that of any dictionary, and extensive training data give them a superior sensitivity to actual usage. All of that promises a considerably better way to predict meaning, but it won't replace judges. Attempting to do so would ignore the model's real limitations, which include their opacity, hallucinatory nature, latent biases, and susceptibility to adversarial attacks by sophisticated parties.

Keeping these limitations in mind, we argued that generative interpretation nevertheless paves an important middle ground between too-cold textualism and too-hot contextualism. The traditional tradeoffs between textualism and contextualism take as a given that our textualist inquiry must depend on dictionaries and that extrinsic evidence is necessarily costly and prone to manipulation. Because generative interpretation is easy to deploy, cheap, and accurate, and because it is not prone to those specific biases, it suggests a workable third way. We argue that, given this technology, parties would prefer courts to ascertain meaning using *some* extrinsic evidence. As such, generative interpretation will become a majoritarian default.

With time, these discussions will spill over to even broader debates about statutory and Constitutional interpretation, originalism and public meaning, and the relative competencies of courts and agencies to reach unbiased, predictable outcomes. We deferred direct discussion of these issues, not least because we're not competent to resolve them. This work, nonetheless, fits into this broader interpretative project of assigning meaning to legal instruments.

We close by offering a different sort of prediction. If, in fact, these models can ascertain party intent to a close-enough approximation, it seems obvious that courts will (and should) use them to make interpretation better. But if that's really true, we wonder why parties would continue to commit to contracts at all? Formal contracting is expensive. Why not, instead, simply write out jointly-held goals at the beginning of the relationship and let models spit out codes of conduct and legal responsibility as problems arise down the line?[229] Or to put it differently, right now, generative AI looks like a promising judicial adjunct. But the future of this technology is more disruptive by far: formal contracts themselves may be made obsolete. Or, at the very least, jurists should consider the marginal value of contracting if the terms themselves are fairly determinable from the parties' goals.

---

[229] *Cf.* Cathy Hwang, *Deal Momentum,* 65 UCLA L. REV. 376, 380 (2018) (describing the use of terms sheets as deal motivators).